\documentclass[]{fairmeta}

\usepackage{amsmath,amsfonts,bm}

\def\eqref#1{equation~\ref{#1}}

\def\1{\bm{1}}

\DeclareMathAlphabet{\mathsfit}{\encodingdefault}{\sfdefault}{m}{sl}
\SetMathAlphabet{\mathsfit}{bold}{\encodingdefault}{\sfdefault}{bx}{n}

\usepackage{hyperref}
\usepackage{url}
\usepackage{booktabs}
\usepackage{amsmath}
\usepackage{amssymb}
\usepackage{amsthm}
\usepackage{algorithm}
\usepackage[noend]{algpseudocode}
\usepackage{graphicx}
\usepackage{wrapfig}
\usepackage[most]{tcolorbox}
\usepackage[table]{xcolor}

\usepackage{enumitem}

\newtheorem{remark}{Remark}

\title{PBPO: Prefix-Block Policy Optimization for Diffusion Language Models}

\author[1,2,*]{Katherine Tieu}
\author[1]{Dongqi Fu} 
\author[1]{Si Zhang}
\author[1]{Hanqing Zeng}
\author[1]{Tianxin Wei}
\author[1]{Chenglin Wei}
\author[1]{Yinglong Xia}
\author[1]{Hong Li}
\author[2]{Jingrui He}

\affiliation[1]{Meta}
\affiliation[2]{University of Illinois Urbana-Champaign}

\contribution[*]{Work done at Meta}

\abstract{
Diffusion Large Language Models (dLLMs) commonly decode responses by proceeding autoregressively across blocks while
iteratively denoising multiple tokens within each block. This structure makes
block-level policy optimization a natural adaptation of autoregressive
policy optimization. However, the exact likelihood of a completed block is
intractable because it marginalizes over the intermediate states and reveal
orders of the denoising process. To realize the block-level formulation, we
construct a prefix-block state and use a single model evaluation to define a
factorized one-step surrogate for the block likelihood. The resulting block
surrogate ratio is a product of its token surrogate ratios, allowing variation
at individual positions to propagate across the entire block before clipping.
We therefore introduce Prefix-Block Policy Optimization (PBPO), which preserves
block-level conditioning but applies surrogate importance weighting and
clipping separately to each token. PBPO further diversifies repeated updates
by resampling corrupted views of the prompt and revealed prefix, a design
motivated by masked-dLLM pre-training. Under our analysis assumptions,
block-product weighting amplifies the variance of gradient contributions
relative to token-local weighting. Across two LLaDA-family initializations,
six benchmarks, and three generation lengths, PBPO achieves competitive
performance against existing dLLM RL methods.
}

\date{\today}

\begin{document}

\maketitle

\section{Introduction}
Reinforcement learning with verifiable rewards has substantially improved the
reasoning capabilities of autoregressive large language models
(LLMs)~\citep{shao2024deepseekmath,guo2025deepseekr1,yu2025dapo}. Masked
diffusion language models (dLLMs) provide a complementary generation paradigm:
rather than predicting strictly from left to right, they iteratively denoise
masked positions and can update multiple tokens in parallel. Recent works such
as LLaDA~\citep{nie2025llada} and LLaDA-1.5~\citep{zhu2025llada15} scale this paradigm to open-ended language generation and
employ block-wise decoding, which combines autoregressive dependencies
across blocks with parallel denoising within each
block. Effectively adapting RL post-training
to this decoding structure remains an important challenge, which lies in likelihood evaluation, as autoregressive policies admit a
unique left-to-right factorization, so each generated token has a tractable
conditional likelihood and importance sampling ratio. In contrast, for a dLLM,
its response likelihood marginalizes over all denoising trajectories
and is therefore intractable to evaluate, preventing a direct adaptation of
autoregressive policy optimization objectives.

Recent work addresses this challenge through different approximations and
reformulations. d1 uses one-step token predictions under a mean-field
approximation~\citep{zhao2025d1}, GDPO estimates a sequence-level diffusion
lower bound~\citep{rojas2025gdpo}, wd1 removes explicit policy ratios in favor
of reward-weighted likelihood optimization~\citep{tang2025wd1}, and DMPO casts
post-training as distribution matching to a reward-modified
target~\citep{zhu2025dmpo}. These approaches primarily
differ in how they approximate the intractable response likelihood or or reformulate the intractable objective. We
study a complementary question induced by block-wise decoding: once a surrogate is chosen, at what granularity should policy change be
measured and clipped?

As block-wise generation proceeds autoregressively across blocks, the
most direct adaptation is to treat each completed block as the optimization
unit conditioned on the previously generated blocks. Its exact conditional
likelihood remains intractable because the block itself is produced by
multi-step denoising. We therefore define a \emph{prefix-block state} that
reveals the completed prefix blocks while masking the current and future
blocks, and approximate the completed block with factorized one-step tokens
denoising from this shared state. This construction yields a tractable surrogate for a
direct block-level policy objective. However, the direct construction also exposes an optimization difficulty. The surrogate
ratio of a block is the product of all token surrogate ratios within that
block, and this single product weights the aggregate contribution of its
tokens. Consequently, a large or small ratio at one position can affect every
token update in the block, with within-block variation compounding before
clipping. Our analysis characterizes this variance amplification under its
stated assumptions, and we observe the corresponding instability when training
the direct block-level formulation.


We introduce \textbf{Prefix-Block Policy Optimization (PBPO)}, which evaluates every token in the current block conditioning on the same prefix-block state, but clips each token’s surrogate ratio independently. When a rollout batch is reused for multiple updates, PBPO also resamples corrupted views of the prompt and revealed prefix. Motivated by masked-dLLM pre-training, this per-update corruption diversifies the conditioning signal without requiring additional computational overhead on rollouts generation. Our main contributions are:
\begin{itemize}[leftmargin=*]
    \item We formulate two policy optimization objectives for dLLMs: a direct block-level objective based on a factorized
    one-step likelihood surrogate, and PBPO, which retains prefix-block
    conditioning while applying clipped surrogate ratios at token level.
    \item We analyze why multiplicative block-level ratios can amplify gradient
    variance under our assumptions and empirically demonstrate that PBPO
    provides more stable optimization than the direct block-level adaptation.
    \item We evaluate PBPO across two LLaDA-family initializations and six
    benchmarks under varied generation lengths and denoising settings, where it
    achieves competitive performance compared to existing dLLM RL methods.
\end{itemize}

\section{Preliminaries}
\subsection{Masked Diffusion Language Models}
\label{sec:prelim_dllm}
Foundational masked diffusion models specialize discrete diffusion to an
absorbing-state corruption process~\citep{shi2024md4,sahoo2024mdlm,ou2025radd}.
Let $\mathcal{V}$ denote the token vocabulary,
$x_0=(x_0^1,\ldots,x_0^L)\in\mathcal{V}^L$ a clean sequence of length $L$, and
$\mathtt{[MASK]}$ the absorbing mask token. At diffusion time $t\in[0,1]$,
the forward process independently masks each token with probability
$1-\alpha_t$, where the noise schedule $\alpha_t\in[0,1]$ decreases from
$\alpha_0=1$ to $\alpha_1=0$, producing a corrupted sequence $x_t$ according to
$
q_{t|0}(x_t\mid x_0)
=
\prod_{k=1}^{L}
\left(
\alpha_t\,\mathbf{1}[x_t^k=x_0^k]
+
(1-\alpha_t)\,\mathbf{1}[x_t^k=\mathtt{[MASK]}]
\right)
$.
As $t$ increases, $x_t$ therefore transitions from clean data to a fully
masked sequence. The reverse process is parameterized by a bidirectional denoising model
$f_\theta$ that predicts the clean token at every masked position. For
absorbing diffusion, the negative evidence lower bound reduces to a weighted
cross-entropy over these predictions~\citep{shi2024md4,sahoo2024mdlm}. Modern dLLMs
scale this formulation using large Transformer models, for instance,
LLaDA~\citep{nie2025llada} adopts the linear schedule $\alpha_t=1-t$, yielding
\begin{equation}
\mathcal{L}_{\mathrm{DCE}}(\theta)
=
-
\mathbb{E}_{
\substack{
x_0\sim p_{\mathrm{data}},\;
t\sim\mathcal{U}[0,1],\\
x_t\sim q_{t|0}(\cdot\mid x_0)
}}
\left[
\frac{1}{t}
\sum_{k=1}^{L}
\mathbf{1}[x_t^k=\mathtt{[MASK]}]
\log f_{\theta}(x_0^k\mid x_t)
\right],
\label{eq:dllm_dce}
\end{equation}

where $p_{\mathrm{data}}$ is the distribution of clean sequences.
Equation~\ref{eq:dllm_dce} is an upper bound on the negative log-likelihood
of the induced generative distribution. For conditional language generation,
let $q$ and $o$ denote the prompt and response, and let $\pi_\theta(o\mid q)$
denote the conditional response distribution induced by the complete reverse process.
Modern dLLMs apply the masking and denoising processes only to $o$: the
response begins fully masked and is iteratively revealed through denoising. Therefore,
the conditional response likelihood is
$
\pi_\theta(o\mid q)
=
\mathbb{E}_{\sigma}
\left[
\prod_{m=1}^{L}
f_\theta
\left(
o^{\sigma_m}
\mid q,o^{\sigma_{<m}}
\right)
\right]
$,
 where $\sigma=(\sigma_1,\ldots,\sigma_L)$ is a token-reveal order, and positions
in $\sigma_{<m}$ are revealed and all remaining response positions are masked. Thus,
$\log\pi_\theta(o\mid q)$ is the log of a marginal over reveal orders, making
its exact evaluation intractable.



\subsection{Reinforcement Learning for Large Language Models}
\label{sec:prelim_rl}
Reinforcement learning with verifiable rewards (RLVR) improves LLM reasoning
by optimizing sequence-level feedback
\citep{ouyang2022training,shao2024deepseekmath,guo2025deepseekr1}. PPO
\citep{schulman2017proximal} stabilizes policy updates through clipped
importance ratios, while GRPO~\citep{shao2024deepseekmath} estimates advantages
from a group of responses and avoids training a separate value model. To be more
specific, given a prompt $q\sim\mathcal{D}$, the behavior policy $\pi_{\mathrm{old}}$
samples $G$ responses
$\{o_i\}_{i=1}^{G}\overset{\mathrm{i.i.d.}}{\sim}
\pi_{\mathrm{old}}(\cdot\mid q)$, and each response receives a reward $R(q,o_i)$.
Following prior work~\citep{liu2025understanding,zhao2025d1}, we use the
unnormalized group-relative advantage
$A_i=R(q,o_i)-\frac{1}{G}\sum_{j=1}^{G}R(q,o_j)$. An
autoregressive policy admits the exact left-to-right factorization
$
\pi_{\theta}(o_i\mid q)
=
\prod_{k=1}^{|o_i|}
\pi_{\theta}
\left(
o_i^k\mid q,o_i^{<k}
\right).
$
Each generated token therefore has a unique conditioning state
$(q,o_i^{<k})$, and its importance sampling ratio is
$
r_i^k(\theta)
=
\frac{
\pi_{\theta}
\left(
o_i^k\mid q,o_i^{<k}
\right)
}{
\pi_{\mathrm{old}}
\left(
o_i^k\mid q,o_i^{<k}
\right)
}.
$
The corresponding clipped objective is
\begin{equation}
J_{\mathrm{RL}}(\theta)
=
\mathbb{E}_{
q\sim\mathcal{D},\\
\{o_i\}_{i=1}^{G}
\sim
\pi_{\mathrm{old}}(\cdot\mid q)
}
\Bigg[
\frac{1}{G}
\sum_{i=1}^{G}
\frac{1}{|o_i|}
\sum_{k=1}^{|o_i|}
\min\Big(
r_i^k(\theta) A_i,
\operatorname{clip}
\left(
r_i^k(\theta),
1-\epsilon_l,
1+\epsilon_r
\right)A_i
\Big)
\Bigg].
\label{eq:ar_rl}
\end{equation}
Clipping limits the contribution of ratios outside the interval
$[1-\epsilon_l,1+\epsilon_r]$. This notation includes symmetric clipping in
PPO and GRPO as well as the asymmetric \textsc{Clip-Higher} rule introduced by
DAPO~\citep{yu2025dapo}.


\textbf{Policy optimization for masked dLLMs.} As shown in Section~\ref{sec:prelim_dllm},
the conditional response likelihood $\pi_\theta(o_i\mid q)$ is intractable, resulting in
policy optimization objectives involving $\pi_\theta(o_i\mid q)$ intractable. d1
addresses this challenge using one-step token predictions with a mean-field
approximation~\citep{zhao2025d1}. The next section addresses the same central
challenge for block-wise generation by constructing a tractable
block-likelihood approximation, which serves as the starting point for PBPO.

\section{Methodology}
\label{sec:method}

We introduce \textbf{Prefix-Block Policy Optimization (PBPO)}, a policy
optimization method for RL post-training of masked diffusion language models
(dLLMs). PBPO is motivated by the block-wise decoding procedure used by
modern dLLMs~\citep{nie2025llada, zhu2025llada15}. Specifically, the procedure
proceeds autoregressively across blocks
and iteratively denoise multiple positions within the current block. Therefore,
the denoised response is therefore naturally
viewed as a sequence of block-level actions, where each block is conditioned on
the previously generated blocks. The most direct adaptation of autoregressive
policy optimization is to replace its token-level action with this
block-level action.





However, realizing this block-level formulation requires evaluating the conditional
likelihood of each completed block, which is intractable. Specifically, as a
block is produced through multi-step
denoising, obtaining its likelihood requires marginalizing over all possible
intermediate masked states and token-reveal orders that can produce the same
completed block. We therefore approximate the block likelihood with a single
forward pass conditioned on a partially masked response, where the previously
generated blocks are unveiled, and the current and future blocks remain
masked. We refer to this conditioning input, as the \emph{prefix-block
state}, which is formalized in Section~\ref{sec:prefix-states}.
This approximation makes the direct block-level policy-optimization formulation
tractable (Section~\ref{sec:block-level}).
Although block-level policy optimization is the natural starting point, its
importance ratio introduces a distinct optimization difficulty: the ratio for
a block is the product of the ratios of all tokens within that block. In
Section~\ref{sec:analysis}, our analyses further show that this multiplicative
structure can amplify the
variance of policy-gradient estimates, and also result in
unstable optimization in practice. Together, these findings motivate the final
PBPO objective, which is elaborated in Section~\ref{sec:pbpo}.
PBPO retains the prefix-block conditioning states used by the
block-level formulation, but decomposes each
block ratio and applies importance weighting and clipping separately to its
tokens. It thereby preserves alignment with block-autoregressive generation
while avoiding a single multiplicative importance weight for an entire block.


\subsection{Prefix-Block States}
\label{sec:prefix-states}
\label{sec:prefix_block}

As motivated above, approximating the likelihood of each block requires a
conditioning input associated with the corresponding block boundary. Let
$\pi_{\theta'}(o^k \mid q,z)$ denote the categorical probability assigned to
the $k$-th token by one forward pass of a masked dLLM with parameters
$\theta'$, conditioned on prompt $q$ and a partially masked response $z$. A
masked position in $z$ is represented by the special token
$\texttt{[MASK]}$. Let $B_l$ denote the predefined block length.
Given a response $o_i$, we divide
it into $B_i=\left \lceil \frac{|o_i|}{B_l} \right \rceil$
consecutive blocks. The $b$-th block spans token positions $(b-1)B_l+1$
through $\min(bB_l,|o_i|)$. For each $b\in\{1,\ldots,B_i\}$, we define the
corresponding \emph{prefix-block state} as
\begin{equation}
    z_i^{<b}
    =
    o_i^{1:(b-1)B_l}
    \oplus
    \underbrace{
        \texttt{[MASK]}\ldots\texttt{[MASK]}
    }_{
        |o_i|-(b-1)B_l
    },
    \label{eq:prefix_block_state}
\end{equation}
where $\oplus$ denotes concatenation. The state $z_i^{<b}$ reveals the
previously generated $(b-1)$ blocks and masks every position in the current
and future blocks. For the first block, no response tokens have yet been
revealed, so its prefix-block state is fully masked:
$
    z_i^{<1}
    =
    \underbrace{
        \texttt{[MASK]}\ldots\texttt{[MASK]}
    }_{|o_i|}.
$

The prefix-block state therefore represents the conditioning input immediately
before block $b$ is denoised. Thus, the
sequence of prefix-block states preserves the autoregressive dependency
across blocks.





\subsection{Block-Level Policy Optimization}
\label{sec:block-level}
\label{sec:block_objective}

Using the prefix-block state $z_i^{<b}$, we approximate the intractable
conditional likelihood of the $b$-th block with a single model
forward pass. As a masked dLLM predicts all positions in the block in
parallel, so we define the surrogate block probability as the factorized product as follows
\begin{equation}
\pi_{*}
\left(
o_i^{(b-1)B_l+1:\min(bB_l,|o_i|)}
\mid q,z_i^{<b}
\right)
=
\exp
\left(
\sum_{k=(b-1)B_l+1}^{\min(bB_l,|o_i|)}
\log
\pi_{*}
\left(
o_i^k
\mid q,z_i^{<b}
\right)
\right),
\label{eq:block_probability}
\end{equation}
where $*$ denotes either the current policy $\pi_\theta$ or the behavior
policy $\pi_{\mathrm{old}}$. The surrogate
retains the block-autoregressive conditioning structure because each block is
evaluated from its corresponding $z_i^{<b}$. Applying the same approximation
to both policies gives the block-level surrogate ratio
\begin{align}
r_i^b
&=
\frac{
\pi_{\theta}
\left(
o_i^{(b-1)B_l+1:\min(bB_l,|o_i|)}
\mid q,z_i^{<b}
\right)
}{
\pi_{\mathrm{old}}
\left(
o_i^{(b-1)B_l+1:\min(bB_l,|o_i|)}
\mid q,z_i^{<b}
\right)
}
=
\prod_{k=(b-1)B_l+1}^{\min(bB_l,|o_i|)}
\frac{
\pi_{\theta}
\left(
o_i^k
\mid q,z_i^{<b}
\right)
}{
\pi_{\mathrm{old}}
\left(
o_i^k
\mid q,z_i^{<b}
\right)
}.
\label{eq:block_ratio}
\end{align}

Using the group-relative sequence-level advantage
$    A_i
    =
    R(q,o_i)
    -
    \frac{1}{G}
    \sum_{j=1}^{G}R(q,o_j)
$, where details for task-specific reward definitions used in our
experiments are provided in Appendix~\ref{app:datasets_rewards},
the block-level PPO-style objective can be formulated as
\begin{equation}
J_{\mathrm{block}}(\theta)
=
\mathbb{E}_{q \sim \mathcal{D},
 \{o_i\}_{i=1}^{G}
\sim
\pi_{\mathrm{old}}(\cdot\mid q)
}
\Bigg[
\frac{1}{G}
\sum_{i=1}^{G}
\frac{1}{B_i}
\sum_{b=1}^{B_i}
\min\Big(r_i^b A_i,\operatorname{clip}
\left(
r_i^b,
1-\epsilon_l,
1+\epsilon_r
\right)A_i
\Big)
\Bigg].
\label{eq:block_objective}
\end{equation}
The next subsection retains its prefix-block conditioning while
revising the granularity of importance weighting.

\subsection{Prefix-Block Policy Optimization}
\label{sec:pbpo}

The block-level objective (Equation~\ref{eq:block_objective}) follows the
generation structure, but its product ratio couples the policy changes of all tokens
within a block. PBPO removes
this coupling while preserving the prefix-block conditioning introduced above.
Specifically, all tokens within a block remain conditioned on the same
$z_i^{<b}$, but each token receives its own importance ratio and clipping
operation. PBPO therefore separates the \emph{conditioning granularity}, which
remains block-level, from the \emph{importance-weighting granularity}, which is
token-level.

For the $t$-th token of response $o_i$, let its block index be $b_i^t = \left\lceil \frac{t}{B_l} \right\rceil$.
PBPO defines the token-level surrogate ratio under the corresponding
prefix-block state as
$
    r_i^t
    =
    \frac{
        \pi_{\theta}
        \left(
            o_i^t
            \mid
            q,z_i^{<b_i^t}
        \right)
    }{
        \pi_{\mathrm{old}}
        \left(
            o_i^t
            \mid
            q,z_i^{<b_i^t}
        \right)
    }
$
Thus, PBPO optimizes the
following objective
\begin{equation}
J_{\mathrm{PBPO}}(\theta)
=
\mathbb{E}_{q \sim \mathcal{D}
\{o_i\}_{i=1}^{G}
\sim
\pi_{\mathrm{old}}(\cdot\mid q)
}
\Bigg[
\frac{1}{G}
\sum_{i=1}^{G}
\frac{1}{|o_i|}
\sum_{t=1}^{|o_i|}
\min\Big(r_i^t A_i,\operatorname{clip}
\left(
r_i^t,
1-\epsilon_l,
1+\epsilon_r
\right)A_i
\Big)
\Bigg].
\label{eq:pbpo}
\end{equation}
Equation~\ref{eq:pbpo} clips each
$r_i^t$ independently, as a policy change at one position therefore affects that
token's surrogate term rather than being multiplied into a shared ratio for
the entire block.
PBPO remains distinct from autoregressive token-level policy optimization
through its conditioning states. An autoregressive model evaluates token
$o_i^t$ from the token prefix $o_i^{<t}$, whereas PBPO evaluates it from the
prefix-block state $z_i^{<b_i^t}$. Tokens in the same block share this state,
while tokens in different blocks condition on different revealed prefixes.
Consequently, token-level importance weighting does not require token-by-token
forward passes, as a single dLLM forward pass produces the probabilities for
all tokens that share a prefix-block state.

\subsubsection{Diversified Conditioning via Per-Update Masking}
\label{sec:masking}

\begin{algorithm}[t]
\caption{Prefix-Block Policy Optimization (PBPO)}
\label{alg:pbpo}
\begin{algorithmic}[1]
\Require Initial masked dLLM policy $\pi_0$, prompt distribution $\mathcal{D}$, reward function $R$, group size $G$, and block length $B_l$, block-diffusion sampling strategy $\mathcal{S}$, inner updates $\mu$, masking probabilities $p_q$ and $p_z$, and clipping coefficients $\epsilon_l$, $\epsilon_r$.
\State Initialize the trainable policy $\pi_\theta \gets \pi_0$
\While{not converged}
    \State Set the behavior policy $\pi_{\mathrm{old}} \gets \pi_\theta$
    \State Sample a prompt $q \sim \mathcal{D}$, and $G$ responses $\{o_i\}_{i=1}^{G} \sim \pi_{\mathrm{old}}(\cdot \mid q)$ using $\mathcal{S}$
    \State Compute $R(q,o_i)$ and $A_i \gets R(q,o_i) - \frac{1}{G}\sum_{j=1}^{G}R(q,o_j), \forall i \in \{1,\ldots,G\}$
    \For{$u=1,\ldots,\mu$}
        \For{$i=1,\ldots,G$}
            \State Construct $\widetilde q_i^{(u)} \gets \mathcal{M}_{p_q}(q)$
            \State Partition $o_i$ into $B_i \gets \lceil |o_i|/B_l \rceil$ consecutive blocks
            \For{$b=1,\ldots,B_i$}
                \State Construct the corrupted prefix-block state $\widetilde z_i^{<b,(u)} \leftarrow$ Equation~\ref{eq:masked_prefix_state}
                \State Compute $r_i^{t,(u)}$ for every token $t$ in block $b \leftarrow$ Equation~\ref{eq:masked_pbpo_ratio}
            \EndFor
        \EndFor
        \Statex \hspace{\algorithmicindent}\(\displaystyle
        \begin{aligned}
        \widehat J_{\mathrm{PBPO}}^{(u)}(\theta)
        ={}& \frac{1}{G}\sum_{i=1}^{G}\frac{1}{|o_i|}\sum_{t=1}^{|o_i|}
        \min\!\Bigl(
        r_i^{t,(u)}A_i, \operatorname{clip}\!\left(r_i^{t,(u)},1-\epsilon_l,1+\epsilon_r\right)A_i
        \Bigr).
        \end{aligned}\)
        \State Update $\theta$ by applying one optimizer step to minimize $-\widehat J_{\mathrm{PBPO}}^{(u)}(\theta)$
    \EndFor
\EndWhile
\State \Return $\pi_\theta$
\end{algorithmic}
\end{algorithm}

Masked dLLMs are pre-trained to predict clean tokens from randomly corrupted
inputs. For example, LLaDA samples a noise level and constructs a corresponding
partially masked sequence during pre-training~\citep{nie2025llada}. This
training paradigm motivates exposing a fixed rollout through different
corrupted conditioning views during policy optimization, with the goal of
diversifying the training signal. We therefore resample the
conditioning context at every update. Following d1~\citep{zhao2025d1}, we
independently resample a random prompt mask. We additionally sample one prefix
masking pattern for each response;
for block $b$, we restrict this pattern to its $(b-1)$ previously generated
blocks. To be more specific, let $\mathcal{M}_{p}(x)$ denote the resulting masked
sequence of $x$, in which each token is independently replaced with
$\texttt{[MASK]}$ with probability $p$.
Suppose $\{o_i\}_{i=1}^{G} \sim \pi_{\mathrm{old}}(\cdot\mid q)$ has been
collected and is reused for $\mu$ policy-update iterations. We use $p_q$ and
$p_z$ to denote the masking probabilities for the prompt and revealed prefix,
respectively. For each response $o_i$ and update iteration
$u\in\{1,\ldots,\mu\}$, each prompt token is independently masked with
probability $p_q$, so the corrupted prompt is constructed as
$
    \widetilde q_i^{(u)}
    =
    \mathcal{M}_{p_q}(q)
$.
Then, we sample one response-wide masking pattern by selecting
each response position independently with probability $p_z$. For obtaining
token's likelihood in block $b$,
we apply only the portion of this pattern corresponding to the revealed prefix
$o_i^{1:(b-1)B_l}$. The corrupted prefix-block state for block $b$ can be
obtained as
\begin{equation}
    \widetilde z_i^{<b,(u)} = \left( \mathcal{M}_{p_z}(o_i) \right)^{1:(b-1)B_l} \oplus \underbrace{\texttt{[MASK]}\ldots\texttt{[MASK]}
}_{
    |o_i|-(b-1)B_l
},
\label{eq:masked_prefix_state}
\end{equation}
which additionally corrupts only the revealed
prefix; the current and future blocks remain fully masked as in the original
prefix-block state $z^{< b}_i$. The resulting PBPO importance ratio at update
iteration $u$ is
\begin{equation}
    r_i^{t,(u)}
    =
    \frac{
        \pi_{\theta}
        \left(
            o_i^t
            \mid
            \widetilde q_i^{(u)},
            \widetilde z_i^{<b_i^t,(u)}
        \right)
    }{
        \pi_{\mathrm{old}}
        \left(
            o_i^t
            \mid
            \widetilde q_i^{(u)},
            \widetilde z_i^{<b_i^t,(u)}
        \right)
    }.
    \label{eq:masked_pbpo_ratio}
\end{equation}
Per-update masking therefore converts one rollout
batch into multiple conditioning views without additional rollout generation.
Empirically, \textit{only applying prompt masking} and \textit{masking both prompt and prefix-block states} consistently
yield better results than clean prompt and clean prefix-block states, as reported in
Table~\ref{tab:pbpo_ablation} in Section~\ref{sec:exp_ablation}.

The complete algorithmic training procedure of PBPO is presented in Algorithm~\ref{alg:pbpo}.


\subsection{Why Token-Level Importance Weighting?}
\label{sec:analysis}

\begin{wrapfigure}{r}{0.5\textwidth} \vspace{-1em} 
\centering
    \vspace{-3mm}
    \includegraphics[width=0.5\linewidth]{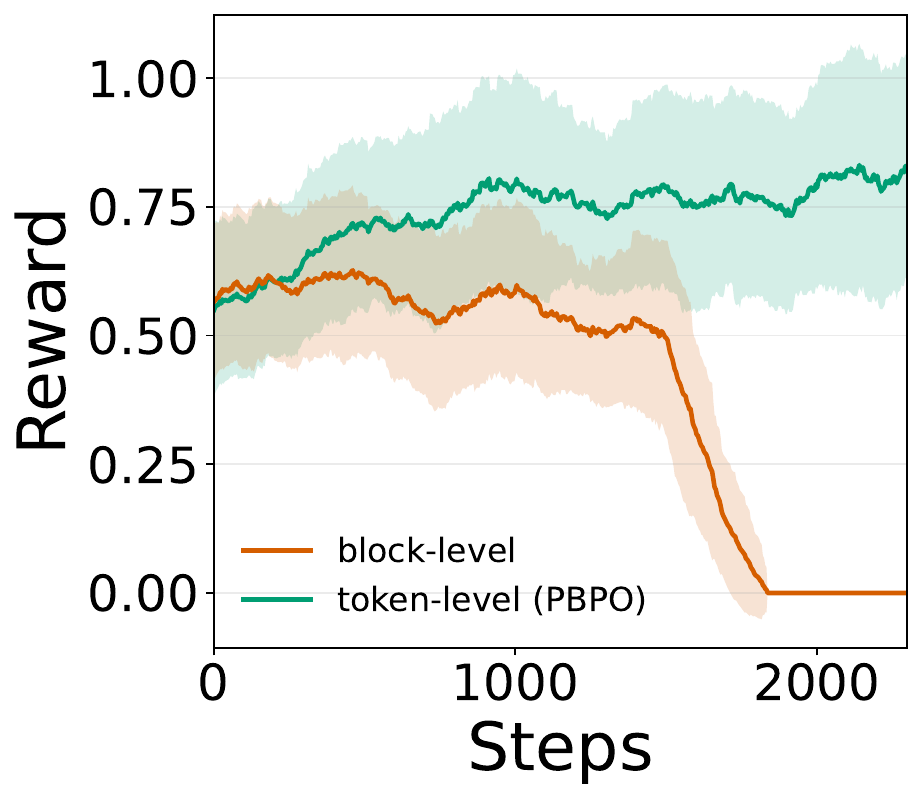}
    \vspace{-10pt}
    \caption{Reward dynamics on MATH500, trained upon LLaDA-8B-Instruct. PBPO demonstrates stable training, while block-level objective collapses early during training.}
    \label{fig:block_collapse}
    \vspace{-4.5mm}
\end{wrapfigure}

The block-level objective (Equation~\ref{eq:block_objective}) and PBPO use the same prefix-block conditioning but
differ in the granularity of importance weighting. We now analyze how this
distinction affects gradient variance. Theorem~\ref{thm:block_noise} formalizes
the comparison (the full proof deferred to
Appendix~\ref{app:theorem-proof}), showing that the multiplicative block ratio couples
the contributions of all tokens in a block: an unusually large or small ratio
at one position can influence the update contributed by the entire block. PBPO
instead confines each importance weight to its corresponding token, avoiding
this within-block propagation. Empirically, we observe the same behavior. In Figure~\ref{fig:block_collapse}, under identical training
configurations, the block-level formulation becomes unstable and collapses
earlier during training, while PBPO demonstrates substantially more stable
optimization. 

\definecolor{thmfill}{HTML}{F0F8FF}   
\definecolor{thmframe}{HTML}{000000}  
\newtheorem{thmbase}{Theorem}

\newtcolorbox{thmbox}[1][]{
  enhanced,
  breakable,
  colback=thmfill,
  colframe=thmframe,
  boxrule=0.9pt,
  arc=6pt,
  left=7pt,
  right=7pt,
  top=4pt,
  bottom=4pt,
  before skip=8pt,
  after skip=8pt,
  #1
}

\newenvironment{boxedthm}[1][]
  {\begin{thmbox}\begin{thmbase}[#1]}
  {\end{thmbase}\end{thmbox}}

\begin{boxedthm}[Blockwise Importance Weighting Produces Noisier Gradient Contributions]
\label{thm:block_noise}
Given a question $q$, and a group of $G$ rollouts
$\{o_i\}_{i=1}^G \sim \pi_{\mathrm{old}}(\cdot \mid q)$.
For any $b$-th block of a rollout $o_i$, let the block-wise gradient contribution of the
$b$-th block of $o_i$, $o_i^{(b-1)B_l+1:bB_l}$, in Equation~\ref{eq:block_objective} be
\begin{equation}
    g_{\mathrm{block}}
    =
    \left[
    \prod_{k=(b-1)B_l+1}^{bB_l}
    \frac{\pi_{\theta}(o_i^k \mid q, z_i^{<b})}
         {\pi_{\mathrm{old}}(o_i^k \mid q, z_i^{<b})}
    \right]
    \left[
    \sum_{k=(b-1)B_l+1}^{bB_l}
    \nabla_{\theta} \log \pi_{\theta}(o_i^k \mid q, z_i^{<b})
    \right],
\end{equation}
and let the gradient contribution of the tokens $o_i^{(b-1)B_l+1:bB_l}$ in
Equation~\ref{eq:pbpo} be
\begin{equation}
    g_{\mathrm{token}}
    =
    \sum_{k=(b-1)B_l+1}^{bB_l}
    \frac{\pi_{\theta}(o_i^k \mid q, z_i^{<b})}
         {\pi_{\mathrm{old}}(o_i^k \mid q, z_i^{<b})}
    \nabla_{\theta} \log \pi_{\theta}(o_i^k \mid q, z_i^{<b}).
\end{equation}
Under the additional assumptions stated in
Appendix~\ref{app:theorem-proof}, we have
\begin{equation}
    \operatorname{Var}\left[g_{\mathrm{block}}\right]
    >
    \operatorname{Var}\left[g_{\mathrm{token}}\right].
\end{equation}
Thus, block-wise importance weighting in Equation~\ref{eq:block_objective} induces noisier
gradient contributions than the token-wise importance weighting used by PBPO.
\end{boxedthm}

\section{Experiments}
\label{sec:experiments}



\subsection{Experimental Settings}
\label{sec:exp_setup}

\textbf{Datasets and evaluation metrics.}
We evaluate PBPO on six benchmarks spanning mathematical reasoning, planning, and coding. For (1) mathematical reasoning,
we train on the GSM8K's training split and MATH dataset~\citep{hendrycks2021math} (without the subset of 500 problems from MATH500~\citep{lightman2023math}) and evaluate on the GSM8K test
set~\citep{cobbe2021gsm8k} and MATH500~\citep{lightman2023math}, respectively.
For planning, we train on three-number TinyZero
Countdown problems~\citep{pan2025tinyzero} and a $4\times4$ Sudoku split, then
evaluate on $256$ held-out Countdown problems and a solution-disjoint Sudoku
set~\citep{arel2025sudoku,wang2025spg}. For coding, we train on
KodCode-Light-RL-10K~\citep{xu2025kodcode} and evaluate on HumanEval's $164$
problems~\citep{chen2021humaneval} and the $257$ tasks in sanitized
MBPP~\citep{austin2021mbpp}. We report accuracy using final-answer matching for
GSM8K and MATH500, valid target-reaching expressions for Countdown, and
complete-puzzle correctness for Sudoku; for HumanEval and MBPP, we report
execution-based \texttt{pass@1}. Further dataset, split, and training reward
details are provided in Appendix~\ref{app:datasets_rewards}.

\textbf{Base models and baselines.}
We initialize RL training from two masked diffusion language models,
\textbf{LLaDA-8B-Instruct}~\citep{nie2025llada} and
\textbf{LLaDA-1.5}~\citep{zhu2025llada15}. We compare PBPO against recent approaches for RL post-training of dLLMs,
including d1~\citep{zhao2025d1}, wd1~\citep{tang2025wd1}, GDPO~\citep{rojas2025gdpo}, and
DMPO~\citep{zhu2025dmpo}.


\textbf{Training setup.}
We perform task-specific RL post-training and train a separate policy for
each benchmark and initialization checkpoint. Following d1~\citep{zhao2025d1},
we use Low-Rank Adaptation (LoRA)~\citep{hu2022lora} for parameter-efficient
optimization. Unless otherwise specified, all training runs share the same
rollout generation process. We use
confidence-based block-wise decoding proposed in LLaDA~\citep{nie2025llada}, which unmasks the two positions with
the highest confidence within the current block at each
denoising step. Complete training configurations, rollout generation details,
and hyperparameters are provided in Appendix~\ref{app:exp_implementation}.

\textbf{Evaluation protocol.}
For each RL run, we evaluate the \emph{final saved checkpoint} after the
prescribed number of optimization steps, with GDPO as the only exception.
As GDPO collapses early during training, we evaluate its last saved
checkpoint immediately before the collapse. Following prior dLLM RL
works~\citep{zhao2025d1,tang2025wd1}, we report results at maximum
completion lengths of $128$, $256$, and $512$. For a maximum generation
length of $N$, we use $N/2$ diffusion
steps and block length $B_l = 32$,
utilizing LLaDA's confidence-based unmasking strategy, as described above. During evaluation, we use temperature
$0.0$ for deterministic decoding.

\subsection{Main Results}
\label{sec:main_results}
\begin{table}[t]
\centering
\caption{Main results across six reasoning and coding benchmarks at generation lengths 128, 256, and 512.
Best result per column within each backbone block is in \textbf{bold}, and the second-best is \underline{underlined}.}
\label{tab:main_exp}
\small
\setlength{\tabcolsep}{4pt}
\resizebox{1.0\linewidth}{!}{%
\begin{tabular}{l*{18}{c}}
\toprule
& \multicolumn{3}{c}{\textbf{GSM8K}}
& \multicolumn{3}{c}{\textbf{MATH500}}
& \multicolumn{3}{c}{\textbf{HumanEval}}
& \multicolumn{3}{c}{\textbf{MBPP}}
& \multicolumn{3}{c}{\textbf{Countdown}}
& \multicolumn{3}{c}{\textbf{Sudoku}} \\
\cmidrule(lr){2-4}
\cmidrule(lr){5-7}
\cmidrule(lr){8-10}
\cmidrule(lr){11-13}
\cmidrule(lr){14-16}
\cmidrule(lr){17-19}
\textbf{Baseline / Seq Len}
& \textbf{128} & \textbf{256} & \textbf{512}
& \textbf{128} & \textbf{256} & \textbf{512}
& \textbf{128} & \textbf{256} & \textbf{512}
& \textbf{128} & \textbf{256} & \textbf{512}
& \textbf{128} & \textbf{256} & \textbf{512}
& \textbf{128} & \textbf{256} & \textbf{512} \\
\midrule
\textit{LLaDA-8B-Instruct}
    & 69.37 & 77.94 & 79.61
    & 26.80 & 32.80 & 34.20
    & 23.17 & 31.71 & 30.49
    & 35.41 & 39.69 & 40.86
    & 19.92 & 17.97 & 17.97
    & 9.33 & 5.74 & 5.12 \\
\hspace{1em} d1
    & 72.55 & 78.01 & 81.20
    & 29.00 & 34.40 & 37.60
    & \underline{29.88} & 35.37 & \textbf{37.80}
    & 41.25 & 42.41 & 41.63
    & 30.08 & 23.44 & 30.08
    & 21.29 & 14.98 & 9.76 \\
\hspace{1em} wd1
    & 73.01 & 80.06 & 81.50
    & \underline{32.80} & 34.80 & 36.20
    & 27.44 & 34.15 & \underline{37.20}
    & 42.80 & 42.02 & 42.80
    & 48.83 & 51.17 & 49.22
    & 24.40 & 24.16 & 22.97 \\
\hspace{1em} GDPO
    & \underline{75.66} & 79.91 & 80.59
    & 31.40 & 34.80 & 37.40
    & 23.17 & \textbf{37.20} & 36.59
    & 42.41 & 41.25 & 42.80
    & \textbf{62.50} & \textbf{73.05} & \textbf{73.05}
    & \underline{28.09} & \underline{28.09} & \underline{26.12} \\
\hspace{1em} DMPO
    & 75.59 & \underline{80.74} & \underline{82.79}
    & 29.80 & \underline{35.40} & \underline{38.20}
    & 24.39 & \underline{35.98} & \textbf{37.80}
    & \underline{43.19} & \underline{43.58} & \underline{43.58}
    & 50.00 & 57.81 & \underline{64.84}
    & 21.15 & 17.13 & 15.50 \\
\rowcolor[HTML]{F0F8FF}[\tabcolsep][\tabcolsep]
\rule[-0.85ex]{0pt}{3.1ex}\hspace{1em} \textbf{PBPO (Ours)}
    & \textbf{76.72} & \textbf{84.08} & \textbf{84.38}
    & \textbf{34.80} & \textbf{36.80} & \textbf{41.40}
    & \textbf{30.49} & \textbf{37.20} & \textbf{37.80}
    & \textbf{43.97} & \textbf{48.25} & \textbf{47.47}
    & \underline{52.73} & \underline{64.06} & 64.45
    & \textbf{34.93} & \textbf{38.13} & \textbf{38.13} \\
\midrule
\textit{LLaDA-1.5}
    & 71.27 & 79.38 & \underline{82.11}
    & 28.40 & 32.20 & \underline{38.40}
    & 21.95 & 30.49 & 34.76
    & 36.19 & 44.36 & 38.91
    & 21.09 & 18.36 & 21.09
    & 13.78 & 10.10 & 8.33 \\
\hspace{1em} d1
    & 74.15 & \underline{81.35} & 80.74
    & 28.00 & \underline{36.20} & 36.00
    & \underline{26.22} & 33.54 & 33.54
    & 40.86 & \textbf{46.69} & 42.02
    & 29.69 & 19.14 & 33.20
    & 21.05 & 15.69 & 11.53 \\
\hspace{1em} wd1
    & 73.09 & 78.77 & 81.12
    & 30.00 & 34.80 & 36.40
    & 23.78 & 35.98 & 31.71
    & 40.08 & 40.47 & 42.41
    & 42.97 & 56.25 & 48.05
    & 24.31 & 23.16 & 22.73 \\
\hspace{1em} GDPO
    & \underline{74.60} & 78.39 & 79.38
    & 30.20 & 34.40 & \underline{38.40}
    & 23.78 & 36.59 & \textbf{35.98}
    & 38.91 & \underline{45.91} & 41.25
    & \underline{51.56} & \underline{67.58} & \textbf{71.88}
    & 25.50 & \underline{26.70} & \underline{24.26} \\
\hspace{1em} DMPO
    & 49.81 & 50.87 & 51.33
    & \underline{31.40} & 35.80 & \textbf{39.00}
    & 25.00 & \underline{39.63} & \underline{35.37}
    & \underline{42.41} & 44.36 & \underline{43.19}
    & 49.61 & 58.98 & \underline{69.92}
    & \underline{26.12} & 24.98 & 23.88 \\
\rowcolor[HTML]{F0F8FF}[\tabcolsep][\tabcolsep]
\rule[-0.85ex]{0pt}{3.1ex}\hspace{1em} \textbf{PBPO (Ours)}
    & \textbf{78.24} & \textbf{83.47} & \textbf{84.61}
    & \textbf{33.40} & \textbf{42.00} & \textbf{39.00}
    & \textbf{31.71} & \textbf{41.46} & \textbf{35.98}
    & \textbf{42.80} & \textbf{46.69} & \textbf{44.36}
    & \textbf{57.03} & \textbf{69.14} & 69.53
    & \textbf{37.75} & \textbf{39.33} & \textbf{36.17} \\
\bottomrule
\end{tabular}%
}
\end{table}

\begin{figure}[t]
    \centering
\includegraphics[width=\linewidth]{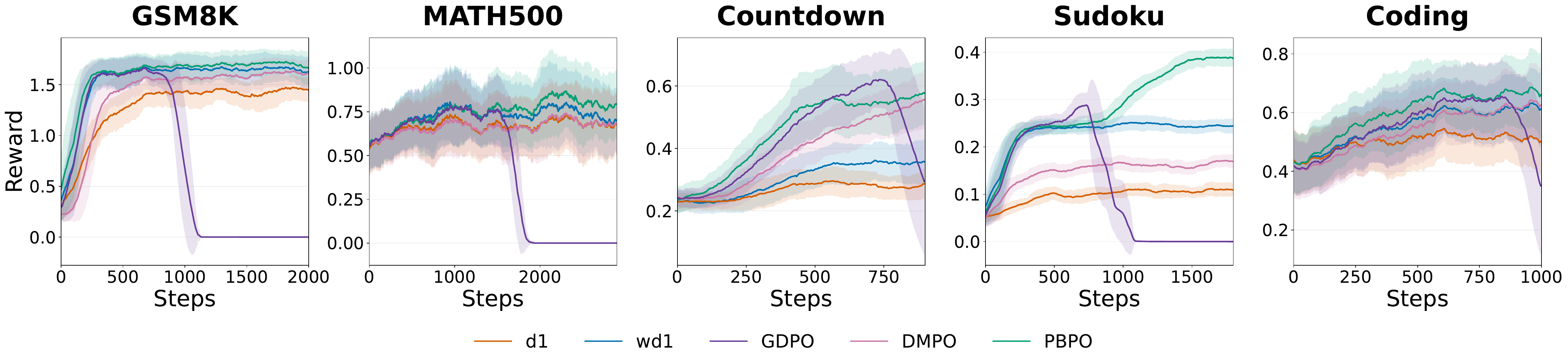}
\vspace{-15pt}
    \caption{Reward dynamics during five RL post-training runs initialized from LLaDA-8B-Instruct. We report the mean and standard deviation over a rolling window of 50 training steps.}
\label{fig:reward_dynamics}
\end{figure}

\begin{figure}[t]
    \centering
\includegraphics[width=\linewidth]{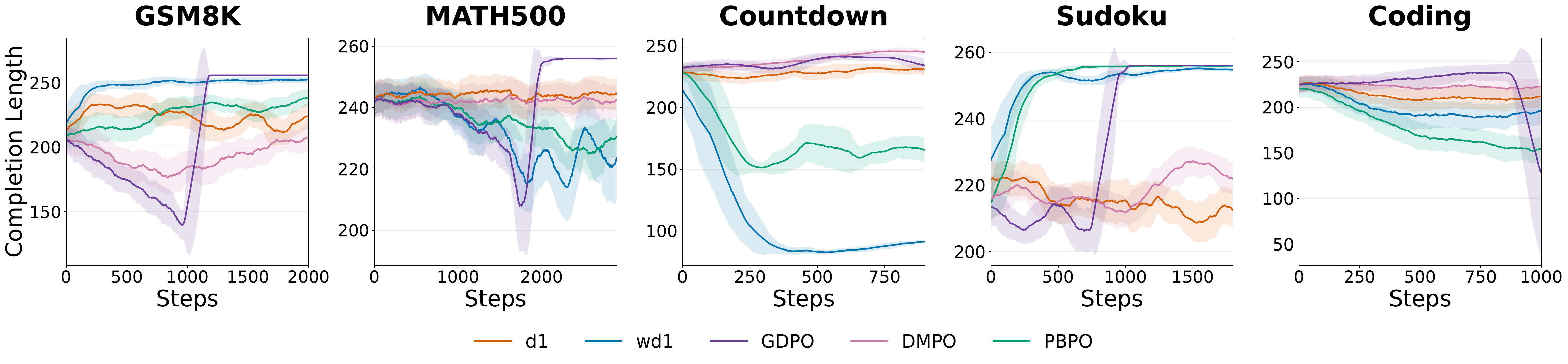}
    \vspace{-15pt}
    \caption{Completion length dynamics during five RL post-training runs initialized from LLaDA-8B-Instruct. We report the mean and standard deviation over a rolling window of 50 training steps.}
\label{fig:completion_dynamics}
\vspace{-2mm}
\end{figure}

\textbf{PBPO achieves competitive performance.}
In Tables~\ref{tab:main_exp}, across all configurations, PBPO outperforms d1, wd1, GDPO, and DMPO in $27$
settings and ties the leading baseline in five more. In the other four
settings, PBPO remains competitive, ranking second or third. This performance
is consistent across both model initializations and task domains: PBPO leads
or ties throughout GSM8K, MATH500, Sudoku, HumanEval, and MBPP, while remaining
competitive on Countdown. Averaged across the three generation lengths, PBPO's
margins over the runner-up on GSM8K, MATH500, Sudoku, HumanEval, and MBPP are
$2.00$, $2.20$, $9.63$, $0.20$, and $3.11$ percentage points, respectively,
with LLaDA-8B-Instruct. With LLaDA-1.5, the corresponding margins are $3.08$,
$2.60$, $12.06$, $2.44$, and $0.52$ percentage points. Qualitative PBPO generations from both model initializations are provided
in Appendix~\ref{app:qualitative_examples}.

\textbf{Training dynamics.}
Figures~\ref{fig:reward_dynamics} and~\ref{fig:llada15_reward_dynamics} show
that PBPO improves steadily and maintains stable optimization throughout
training with both model initializations. It finishes with the highest reward
across all five
training runs, complementing its competitive final evaluation performance in
Table~\ref{tab:main_exp}.

\textbf{Effective utilization of the generation budget.}
Figures~\ref{fig:completion_dynamics} and~\ref{fig:llada15_completion_dynamics}
show that PBPO couples improving reward with effective use of the generation
budget. Although PBPO converges to the highest completion-length regime
alongside wd1 and GDPO on Sudoku, it achieves substantially higher reward and
evaluation accuracy than both. Tables~\ref{tab:main_exp_effective_seq_len}
and~\ref{tab:code_avg_tokens} reinforce this efficiency at evaluation time:
PBPO pairs strong accuracy with compact effective responses, ranking as the
shortest or second-shortest method in $29$ configurations. On the coding
benchmarks, it produces the shortest responses in $11$ of $12$ settings and
the second-shortest in the remaining one while attaining the highest
\texttt{pass@1} throughout. These results show that PBPO yields strong task
performance with effective generation.

\textbf{Robustness across various denoising configurations.}
We further vary the block length $B_l$ and evaluate both confidence-based and
random token-selection strategies, which were proposed in
LLaDA~\citep{nie2025llada}. Complete configuration details
and empirical results are provided in
Appendix~\ref{app:alternative_eval_settings}.
Across the $30$ combinations of six benchmarks and five denoising
configurations, PBPO achieves the best result in $20$, ties the leading
baseline in two, ranks second in five, and ranks third in three.
PBPO achieves the best result on GSM8K and Sudoku under every configuration,
while the two confidence-based block-length variants nearly
preserve its default mean accuracy across all six benchmarks ($53.22\%$ versus
$53.68\%$). Under random token selection, PBPO ranks first or second in $17$
of $18$ comparisons. Its effective sequence length also ranks first or second
in $20$ configurations, demonstrating robust task performance together with
effective generation across various decoding configurations.
\subsection{Ablation Study}
\label{sec:exp_ablation}
We next examine the effect of the conditioning diversification strategy
introduced in Section~\ref{sec:masking}. We compare three ablations:
(1) PBPO with a clean prompt and clean prefix;
(2) PBPO with per-update prompt masking but a clean prefix; and
(3) PBPO with per-update masking of both the prompt and revealed prefix.
For the masked variants, each eligible token is independently replaced by
$\texttt{[MASK]}$ with probability $0.15$. We conduct this ablation on MATH500 and Countdown using LLaDA-1.5 as the
initialization checkpoint.


\begin{table}[t]
\centering
\caption{Ablation across three PBPO variants on Countdown and MATH500 at generation lengths 128, 256, and 512, with LLaDA-1.5 as the base model. Best result per column is in \textbf{bold}, and second-best result is highlighted with \underline{underline}.}
\label{tab:pbpo_ablation}
\small
\setlength{\tabcolsep}{4pt}
\resizebox{0.8\linewidth}{!}{%
\begin{tabular}{lcccccc}
\toprule
& \multicolumn{3}{c}{\textbf{Countdown}}
& \multicolumn{3}{c}{\textbf{MATH500}} \\
\cmidrule(lr){2-4} \cmidrule(lr){5-7}
\textbf{Baseline / Seq Len}
& \textbf{128} & \textbf{256} & \textbf{512}
& \textbf{128} & \textbf{256} & \textbf{512} \\
\midrule
PBPO + clean prompt + clean prefix
    & 46.09 & 47.27 & 51.56
    & 30.80 & 32.60 & 36.20 \\
PBPO + prompt masking + clean prefix
    & \underline{47.27} & \underline{57.81} & \underline{62.89}
    & \textbf{33.40} & \textbf{42.00} & \textbf{39.00} \\
PBPO + prompt masking + prefix masking
    & \textbf{57.03} & \textbf{69.14} & \textbf{69.53}
    & \underline{31.00} & \underline{35.60} & \underline{38.20} \\
\bottomrule
\end{tabular}%
}
\vspace{-4mm}
\end{table}

As shown in Table~\ref{tab:pbpo_ablation}, introducing
per-update corruption consistently improves over using a clean prompt and
clean prefix across both benchmarks and all three generation lengths. This
supports the motivation in Section~\ref{sec:masking}, as exposing the dLLM to
multiple corrupted conditioning views provides a more effective way to reuse
a fixed rollout batch during policy optimization. The effect of additionally masking the revealed prefix is task-dependent.
On Countdown, prompt-and-prefix masking substantially outperforms prompt-only
masking, improving accuracy by $9.76$, $11.33$, and $6.64$ percentage
points at generation lengths $128$, $256$, and $512$, respectively.
On MATH500, prompt-only masking performs better than additionally masking the
prefix, although both masked variants outperform the fully clean
configuration. These results suggest that conditioning diversification is
consistently beneficial, while the preferred amount of corruption in the
revealed prefix can depend on the reasoning task.

\section{Related Works}
\textbf{Masked diffusion language models.}
Masked diffusion language models specialize discrete diffusion to
absorbing-state corruption, with training objectives that reduce to weighted
denoising losses over masked
positions~\citep{shi2024md4,sahoo2024mdlm}. RADD further connects absorbing
diffusion to clean-data conditional prediction and any-order autoregressive
modeling~\citep{ou2025radd}. Building on these foundations, LLaDA and
LLaDA-1.5 scale masked diffusion to large language models and use block-wise
decoding to combine parallel denoising with dependencies across completed
blocks~\citep{nie2025llada,zhu2025llada15}. PBPO focuses on RL post-training
under this block-wise generation structure.

\textbf{Policy optimization for masked dLLMs.}
Because exact response likelihoods are intractable, existing methods adopt
different optimization surrogates. d1 applies group-relative policy
optimization using one-step token predictions under a mean-field
approximation~\citep{zhao2025d1}. wd1 instead avoids explicit importance ratios
and optimizes a reward-weighted token likelihood~\citep{tang2025wd1}. GDPO
constructs a sequence-level ratio from a diffusion evidence lower bound and
uses structured estimation to manage its cost and
variance~\citep{rojas2025gdpo}, whereas DMPO replaces clipped ratio optimization
with matching to a reward-modified target distribution~\citep{zhu2025dmpo}. PBPO retains clipped surrogate ratios and addresses the
distinct question of how these ratios should be assigned within the
block-wise structure.
\vspace{-2mm}

\section{Conclusion}
We propose Prefix-Block Policy Optimization (PBPO) for RL post-training of dLLMs. Starting from the  block-level
adaptation of autoregressive policy optimization, PBPO approximates
intractable completed-block likelihoods with factorized one-step predictions
from a prefix-block state. Whereas the block-level objective multiplies the
token surrogate ratios into one shared weight, PBPO retains the same
prefix-block conditioning while applying surrogate importance weighting and
clipping separately to each token, yielding more stable training. Across two LLaDA-family initializations, six
benchmarks, and varied denoising configurations, PBPO achieves competitive
performance against existing dLLM RL methods while maintaining stable reward
improvement. PBPO still relies on a one-step approximation to the exact
multi-step denoising likelihood; extending the formulation to broader dLLM
families and decoding procedures is therefore an important direction for
future work.

\clearpage
\newpage
\bibliographystyle{assets/plainnat}
\bibliography{paper}

\clearpage
\newpage
\beginappendix

\section{Additional Experimental Details}
\label{app:exp_details}

\subsection{Datasets and Reward Functions}
\label{app:datasets_rewards}

We follow the task-specific verifiable rewards used in prior dLLM RL
work~\citep{zhao2025d1,tang2025wd1}, while specifying the exact
parsing and scoring rules used in our implementation. All reward functions
operate on the decoded response string $o$. When a task uses $M$ reward
components, the sequence-level reward is their equal-weight sum,
\begin{equation}
    R(q,o)=\sum_{m=1}^{M} R_m(q,o),
    \label{eq:task_reward_sum}
\end{equation}
where $R_m$ denotes the $m$-th task-specific component. Countdown and Sudoku
use a single component. PBPO subsequently converts $R(q,o)$ into the
group-relative advantage in Section~\ref{sec:prelim_rl}; no additional reward
normalization is applied.

\paragraph{GSM8K.}
We perform RL training on the standard GSM8K training split and evaluate on
the held-out test split. Let $a(o)$ be the whitespace-trimmed substring after
the final \texttt{<answer>} tag and before the next \texttt{</answer>} tag. If
an opening answer tag is absent, the parser falls back to the response content
preceding a closing answer tag, if present. The ground-truth answer $y$ is the
trimmed substring following the \texttt{\#\#\#\#} delimiter in the GSM8K
reference solution. The training reward is
\begin{equation}
    R_{\mathrm{GSM8K}}(q,o)
    =R_{\mathrm{XML}}(o)+R_{\mathrm{soft}}(o)
    +R_{\mathrm{strict}}(o)+R_{\mathrm{int}}(o)
    +R_{\mathrm{correct}}(o,y),
    \label{eq:gsm8k_reward}
\end{equation}
with the following five components:
\begin{itemize}
    \item \textbf{XML structure.} $R_{\mathrm{XML}}$ adds $0.125$ when each
    of the following literal delimiters occurs exactly once:
    \texttt{<reasoning>\textbackslash n},
    \texttt{\textbackslash n</reasoning>\textbackslash n},
    \texttt{\textbackslash n<answer>\textbackslash n}, and
    \texttt{\textbackslash n</answer>}. The trailing-content adjustment
    follows the implementation literally. If
    \texttt{\textbackslash n<answer>\textbackslash n} occurs exactly once, it
    subtracts $0.001$ times the number of characters returned after splitting
    on \texttt{\textbackslash n</answer>\textbackslash n}; if that closing
    delimiter is absent, this operation returns the full response. A second
    check, activated when \texttt{\textbackslash n</answer>} occurs exactly
    once, subtracts $0.001(L-1)$, where $L$ is the length of the suffix after
    that delimiter. This operation does not verify that the omitted character
    is a newline: it adds $0.001$ when the suffix is empty and otherwise ignores
    one trailing character. The two adjustments are added when both checks
    apply.
    \item \textbf{Soft format.} $R_{\mathrm{soft}}=0.5$ when the response
    matches the code-like pattern 
    \newline
    \texttt{<reasoning>...</reasoning>\textbackslash s*<answer>...</answer>}
    from the beginning of the string, and is zero otherwise. Because the
    regular expression does not enable multiline matching, the content inside
    each tag pair must lie on a single line.
    \item \textbf{Strict format.} $R_{\mathrm{strict}}=0.5$ only for the
    exact layout
    \texttt{<reasoning>\textbackslash n...\textbackslash n</reasoning>}
    followed by
    \texttt{\textbackslash n<answer>\textbackslash n...\textbackslash n</answer>\textbackslash n};
    otherwise, it is zero. Here, each ellipsis again matches only single-line
    content. Consequently, the soft and strict format indicators cannot both
    be active for the same response under these regular expressions.
    \item \textbf{Integer answer.} $R_{\mathrm{int}}=0.5$ when every
    character of $a(o)$ is a decimal digit, and is zero otherwise.
    \item \textbf{Answer correctness.}
    $R_{\mathrm{correct}}(o,y)=2\,\mathbf{1}[a(o)=y]$, where equality is an
    exact string comparison after trimming surrounding whitespace.
\end{itemize}
Because the soft and strict patterns are mutually exclusive, the largest
executed reward is $3.5$, attained by a correct integer answer satisfying the
strict layout and all four XML checks. Trailing content can further reduce the
XML component.
Evaluation reports the percentage of examples for which the extracted final
answer exactly matches the ground-truth answer.

\paragraph{MATH500.}
For RL training, we load the training split of the MATH500 dataset and evaluate
on its held-out split. We first extract the final \texttt{\textbackslash
boxed\{\}} expression from the response, or the final
\texttt{\textbackslash fbox\{\}} expression when no
\texttt{\textbackslash boxed} expression occurs; if neither is found, the full
response is retained as the candidate. We remove the outer box and apply
the same procedure to the reference solution. The candidate and reference are
then normalized using a deterministic string procedure. Specifically, the
normalizer removes newline characters and literal spaces, inverse spacing
\texttt{\textbackslash !}, \texttt{\textbackslash left} and
\texttt{\textbackslash right}, degree markers, escaped dollar and percentage
symbols, and all content beginning at the exact substring
\texttt{\textbackslash text\{ } when that substring occurs once. If it
occurs multiple times, normalization fails and the comparator falls back to
raw-string equality. The procedure also canonicalizes doubled backslashes,
\texttt{\textbackslash tfrac}, and \texttt{\textbackslash dfrac}; repairs
simple fraction and square-root syntax; converts a simple $a/b$ expression and
$0.5$ to fractional notation; normalizes leading decimals; and drops a
variable name of at most two characters preceding a single equality. Let
$\mathcal{N}(o)$ and
$\mathcal{N}(y)$ denote the resulting normalized candidate and reference
strings, respectively.

For the format component, let $h_{\mathrm{ans}}(o)$ indicate that the literal
opening tag \texttt{<answer>} occurs; a closing tag is not required by the
implemented split operation. Moreover, let $h_{\mathrm{box}}(o)$ indicate the
literal control-character sequence tested by the implementation, within the
extracted answer content when $h_{\mathrm{ans}}(o)=1$ and within the full
response otherwise. This sequence is \texttt{U+0008oxed} (a backspace followed
by \texttt{oxed}), rather than the printed LaTeX command
\texttt{\textbackslash boxed}; consequently, $h_{\mathrm{box}}(o)=0$ for
ordinary model outputs containing
\texttt{\textbackslash boxed}. The executed reward is
\begin{equation}
\begin{aligned}
    R_{\mathrm{MATH}}(q,o)
    ={}&2\,\mathbf{1}\!\left[\mathcal{N}(o)=\mathcal{N}(y)\right]
    +0.25+0.5h_{\mathrm{ans}}(o)+0.25h_{\mathrm{box}}(o).
\end{aligned}
\label{eq:math_reward}
\end{equation}
Accordingly, correctness is determined by equality after deterministic string
normalization rather than by a computer-algebra solver. For ordinary decoded
text, the format component is $0.75$ when \texttt{<answer>} is present and
$0.25$ otherwise, giving a maximum attainable reward of $2.75$; the literal
program maximum is $3.0$ if the control-character condition is met. Evaluation
applies the same normalized-answer comparison.

\paragraph{Countdown.}
We use the training split of the Countdown dataset from
TinyZero~\citep{pan2025tinyzero}, restricted to problems containing three input
numbers. After shuffling with the experiment seed, we reserve the final $500$
examples and train on the remainder. Following d1~\citep{zhao2025d1}, we evaluate on a fixed set of $256$
synthetically generated three-number Countdown problems.
The parser extracts the final substring enclosed by
\texttt{<answer>} and \texttt{</answer>}. Let $e(o)$ denote this expression,
$T\in\mathbb{R}$ the target, and $\mathcal{N}_q$ the multiset of three input
numbers. The verifier extracts every maximal digit sequence in $e(o)$, converts
each sequence to an integer, and requires the resulting multiset to equal
$\mathcal{N}_q$. It then evaluates expressions containing only digits,
whitespace, decimal points, parentheses, and the operators $+$, $-$, $*$, and
$/$ in an environment without Python built-ins. The reward used in our runs is
\begin{equation}
R_{\mathrm{Countdown}}(q,o)=
\begin{cases}
1, & \substack{\text{if an expression is extracted, uses exactly }\mathcal{N}_q,\\
                 \text{evaluates successfully, and differs from }T\text{ by less than }10^{-5},}\\
0, & \text{if no answer expression can be extracted},\\
0.1, & \text{otherwise}.
\end{cases}
\label{eq:countdown_reward}
\end{equation}
Thus, the $0.1$ fallback also covers an extracted expression with an invalid
number multiset, unsupported characters, an evaluation error, or an incorrect
value; it is not restricted to well-formed expressions that merely miss the
target.

\paragraph{Sudoku.}
We use the $4\times4$ Sudoku dataset generated by
Arel~\citep{arel2025sudoku}, whose original training set contains one million
unique puzzles spanning all $288$ possible $4\times4$ solution grids.
Following~\citep{wang2025spg}, we construct a solution-disjoint
train--test split to prevent leakage through memorization of complete
solutions. Specifically, we randomly select $200$ solution grids and include
all puzzles associated with them in the training pool, resulting in $694{,}006$
puzzles. The loader shuffles this pool with the experiment seed, reserves its
final $500$ examples, and therefore uses $693{,}506$ puzzles for RL training.
The test set contains $256$ puzzles sampled from the remaining $88$ solution
grids, ensuring that no test solution appears in the RL training set.

For a puzzle string $p\in\{0,1,2,3,4\}^{16}$, let $g\in\{1,2,3,4\}^{16}$
denote its ground-truth solution and
$\mathcal{E}(p)=\{j\in\{1,\ldots,16\}:p_j=0\}$ the indices of originally
empty cells. The parser takes the final answer-tag span, removes every
non-digit character, and obtains a candidate digit string $s(o)$. Candidates
shorter than $16$ digits are right-padded with zeros, whereas candidates
longer than $16$ digits are truncated; denote the resulting length-$16$
string by $\bar{s}(o)$. If no nonempty answer can be extracted, the reward is
zero. Otherwise,
\begin{equation}
    R_{\mathrm{Sudoku}}(q,o)
    =
    \frac{1}{|\mathcal{E}(p)|}
    \sum_{j\in\mathcal{E}(p)}
    \mathbf{1}\!\left[\bar{s}_j(o)=g_j\right].
    \label{eq:sudoku_reward}
\end{equation}
Only originally empty cells contribute: copied clues receive no credit, and
there is no separate reward for satisfying row, column, or subgrid constraints.
Evaluation is stricter than the dense training reward and reports the fraction
of puzzles for which the complete solution is correct.

\paragraph{Code generation.}
We perform coding RL on the training portion of
KodCode-Light-RL-10K~\citep{xu2025kodcode}. Following a deterministic split
with seed $42$, we use $90\%$ of its training split; each example provides a
coding problem, a reference solution, and synthetic unit tests. The coding
reward combines the XML-structure component defined for GSM8K with a
code-specific component that evaluates both output placement and functional
correctness.

More precisely, let $c(o)$ denote the program extracted from the first literal
code block matching
\texttt{```python\textbackslash n...```} in response $o$. Let
$h_{\mathrm{place}}(o)$ indicate that this entire code block occurs inside the
first \texttt{<answer>...</answer>} span. For a valid KodCode test
specification, let $\mathcal{T}=\{\tau_j\}_{j=1}^{J}$ denote the $J$ unit tests
associated with the problem, and let
$\operatorname{Pass}(c(o),\tau_j)\in\{0,1\}$ indicate whether executing
$c(o)$ with test $\tau_j$ terminates successfully. The resulting reward is
\begin{equation}
\begin{aligned}
R_{\mathrm{Code}}(q,o)
={}&R_{\mathrm{XML}}(o)
+0.5\,h_{\mathrm{place}}(o)
+\frac{1}{J}\sum_{j=1}^{J}\operatorname{Pass}(c(o),\tau_j),
\end{aligned}
\label{eq:coding_reward}
\end{equation}
where the test-pass fraction is defined as zero when no executable test is
available. Thus, a correctly formatted response whose program passes every
test receives a maximum reward of $2.0$: up to $0.5$ from XML structure, $0.5$
from placing the Python code block inside the answer tags, and $1.0$ from
functional correctness.

Before execution, the verifier identifies the expected function name from a
literal statement of the form \texttt{from solution import <name>} or,
otherwise, from the first assertion that invokes a function. It extracts the
first function defined by the generated program and renames that definition to
the expected name when necessary. The verifier then rejects programs
containing any blocked module or keyword from
\texttt{os}, \texttt{sys}, \texttt{subprocess}, \texttt{shutil},
\texttt{socket}, \texttt{psutil}, \texttt{ctypes}, \texttt{pathlib},
\texttt{builtins}, and \texttt{\_\_import\_\_}. Unsafe programs and outputs
without a function definition receive no functional-correctness credit,
although a correctly placed code block retains its $0.5$ placement reward.
Finally, each unit test is executed in a separate worker process with a
$1$-GB memory limit and a one-second deadline; exceptions, timeouts, and
missing test results are counted as failures. During evaluation on HumanEval
and MBPP, we instead report the stricter execution-based \texttt{pass@1}
metric described in Section~\ref{sec:exp_setup}.

\subsection{Training and Implementation Details}
\label{app:exp_implementation}

Here, we provide the complete rollout, optimization, and implementation
configurations required for reproducibility.

\paragraph{Rollout generation.}
For each prompt, we sample $G=6$ responses with maximum completion length
$256$ and temperature $0.9$. Following LLaDA~\citep{nie2025llada} and prior
dLLM RL work~\citep{zhao2025d1,tang2025wd1}, generation proceeds
block-by-block with block length $B_l=32$ over $128$ denoising steps. At each
step, the two currently masked positions with the highest predicted-token
confidence within the active block are revealed, while all remaining positions
in that block stay masked. Generation advances to the next block after the
active block is fully revealed. Each resulting rollout batch is reused for
$\mu=4$ policy-update iterations.

We use a per-device training batch size of $6$ with $2$ gradient accumulation
steps. Every training run is conducted on $8$ NVIDIA H200 GPUs.

Following d1~\citep{zhao2025d1}, we use LoRA with rank $r=128$, scaling
factor $\alpha=64$, and dropout $0.05$. LoRA adapters are applied to the
query, key, value, and output projections in the attention layers, as well
as the up, down, and gate projections in the feed-forward layers. We optimize
the LoRA parameters using AdamW with $\beta_1=0.9$, $\beta_2=0.99$, weight
decay $0.1$, learning rate $3\times10^{-6}$, and maximum gradient norm $0.2$.
The learning rate follows a \texttt{constant\_with\_warmup} schedule with warmup ratio
$10^{-4}$.

For PBPO with per-update masking, the clean prompt, clean revealed prefixes,
generated responses, and behavior policy $\pi_{\mathrm{old}}$ remain fixed
while a rollout batch is reused. At each policy-update iteration, a fresh
masking pattern is independently sampled and applied to the same clean prompt
and prefix sequences, as described in Section~\ref{sec:masking}.

The optimization, batching, rollout, and LoRA settings are shared across all
experiments. Tables~\ref{tab:training_hparams_llada8b}
and~\ref{tab:training_hparams_llada15} report the dataset-specific
hyperparameters separately for policies initialized from LLaDA-8B-Instruct
and LLaDA-1.5, respectively.

\begin{table}[t]
\centering
\small
\caption{Dataset-specific PBPO training hyperparameters for policies initialized from LLaDA-8B-Instruct.}
\label{tab:training_hparams_llada8b}
\resizebox{\linewidth}{!}{%
\begin{tabular}{lccccc}
\toprule
\textbf{Hyperparameter}
& \textbf{GSM8K}
& \textbf{MATH500}
& \textbf{Countdown}
& \textbf{Sudoku}
& \textbf{KodCode-Light-RL-10K} \\
\midrule
Training steps
    & 2,000 & 2,900 & 900 & 1,800 & 1,000 \\
Prompt masking probability $p_q$
    & 0.15 & 0.15 & 0.15 & 0.15 & 0.15 \\
Prefix masking probability $p_z$
    & 0 & 0 & 0.15 & 0 & 0.15 \\
$\epsilon_l$
    & 0.5 & 0.2 & 0.2 & 0.2 & 0.2 \\
$\epsilon_r$
    & 0.5 & 0.28 & 0.28 & 0.28 & 0.28 \\
\bottomrule
\end{tabular}%
}
\end{table}

\begin{table}[t]
\centering
\small
\caption{Dataset-specific PBPO training hyperparameters for policies initialized from LLaDA-1.5.}
\label{tab:training_hparams_llada15}
\resizebox{\linewidth}{!}{%
\begin{tabular}{lccccc}
\toprule
\textbf{Hyperparameter}
& \textbf{GSM8K}
& \textbf{MATH500}
& \textbf{Countdown}
& \textbf{Sudoku}
& \textbf{KodCode-Light-RL-10K} \\
\midrule
Training steps
    & 2,200 & 2,400 & 1,000 & 1,900 & 1,000 \\
Prompt masking probability $p_q$
    & 0.15 & 0.15 & 0.15 & 0.15 & 0.15 \\
Prefix masking probability $p_z$
    & 0 & 0 & 0.15 & 0 & 0.15 \\
$\epsilon_l$
    & 0.2 & 0.2 & 0.2 & 0.2 & 0.2 \\
$\epsilon_r$
    & 0.28 & 0.28 & 0.28 & 0.28 & 0.28 \\
\bottomrule
\end{tabular}%
}
\end{table}





\section{Additional Empirical Results}

\subsection{Training Dynamics with LLaDA-1.5}

\begin{figure}[t]
    \centering
    \includegraphics[width=\linewidth]{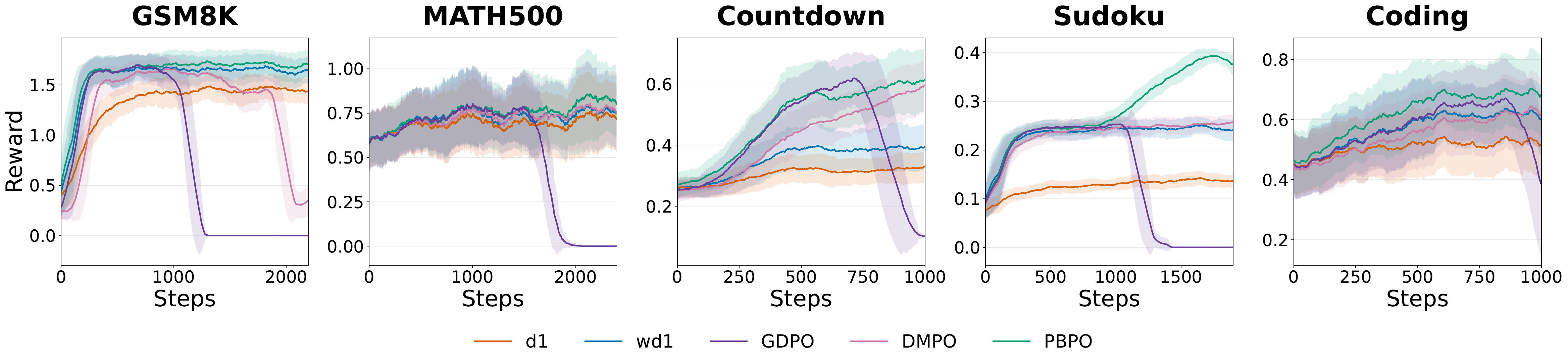}
    \caption{Reward dynamics during five RL post-training runs initialized from LLaDA-1.5. We report the mean and standard deviation over a rolling window of 50 training steps.}
    \label{fig:llada15_reward_dynamics}
\end{figure}

\begin{figure}[t]
    \centering
    \includegraphics[width=\linewidth]{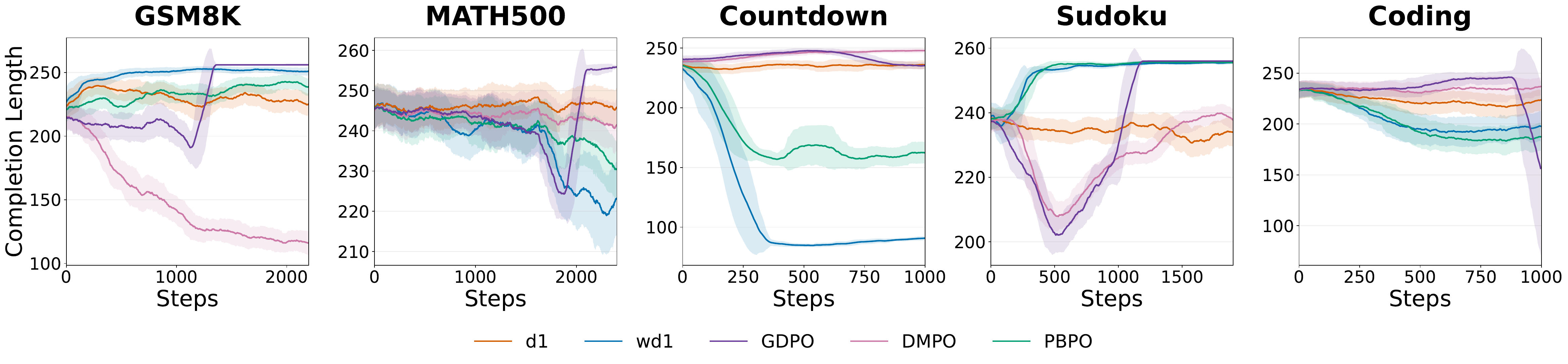}
    \caption{Completion length dynamics during five RL post-training runs initialized from LLaDA-1.5. We report the mean and standard deviation over a rolling window of 50 training steps.}
    \label{fig:llada15_completion_dynamics}
\end{figure}

Figures~\ref{fig:llada15_reward_dynamics}
and~\ref{fig:llada15_completion_dynamics} provide the reward and completion
length trajectories for the five training runs when PBPO and the baselines are
initialized from LLaDA-1.5.

\subsection{Effective Sequence Length Across Benchmarks}

\paragraph{Reasoning benchmarks.}
Table~\ref{tab:main_exp_effective_seq_len} reports the average effective
sequence length of the checkpoints evaluated in Table~\ref{tab:main_exp}
across the four reasoning benchmarks.

\begin{table}[t]
\centering
\caption{Effective sequence length for each baseline across four reasoning benchmarks at generation
lengths 128, 256, and 512.}
\label{tab:main_exp_effective_seq_len}
\small
\setlength{\tabcolsep}{4pt}
\resizebox{1.0\linewidth}{!}{%
\begin{tabular}{lcccccccccccc}
\toprule
& \multicolumn{3}{c}{\textbf{GSM8K}}
& \multicolumn{3}{c}{\textbf{MATH500}}
& \multicolumn{3}{c}{\textbf{Countdown}}
& \multicolumn{3}{c}{\textbf{Sudoku}} \\
\cmidrule(lr){2-4}
\cmidrule(lr){5-7}
\cmidrule(lr){8-10}
\cmidrule(lr){11-13}
\textbf{Baseline / Seq Len}
& \textbf{128} & \textbf{256} & \textbf{512}
& \textbf{128} & \textbf{256} & \textbf{512}
& \textbf{128} & \textbf{256} & \textbf{512}
& \textbf{128} & \textbf{256} & \textbf{512} \\
\midrule
\textit{LLaDA-8B-Instruct}
    & 114.31 & 211.21 & 256.14
    & 121.62 & 234.50 & 399.00
    & 110.49 & 212.89 & 409.96
    & 115.42 & 216.92 & 432.95 \\
\hspace{1em} d1
    & 115.85 & 217.13 & 275.15
    & 122.05 & 234.47 & 392.75
    & 107.49 & 209.75 & 397.20
    & 115.54 & 225.42 & 435.74 \\
\hspace{1em} wd1
    & 114.25 & 226.42 & 326.66
    & 120.21 & 209.63 & 319.50
    & 76.34 & 80.60 & 85.30
    & 111.13 & 236.70 & 465.87 \\
\hspace{1em} GDPO
    & 109.46 & 192.20 & 204.85
    & 120.45 & 222.81 & 367.36
    & 114.25 & 213.27 & 351.79
    & 114.17 & 213.92 & 355.32 \\
\hspace{1em} DMPO
    & 114.22 & 213.28 & 258.28
    & 121.78 & 234.64 & 395.15
    & 110.21 & 212.17 & 364.04
    & 114.11 & 228.31 & 243.07 \\
\rowcolor[HTML]{F0F8FF}[\tabcolsep][\tabcolsep]
\rule[-0.85ex]{0pt}{3.1ex}\hspace{1em} PBPO (Ours)
    & 113.78 & 221.08 & 281.70
    & 121.62 & 226.61 & 359.10
    & 103.37 & 134.61 & 159.16
    & 102.94 & 164.98 & 244.54 \\
\midrule
\textit{LLaDA-1.5}
    & 115.15 & 214.51 & 263.16
    & 123.23 & 237.43 & 405.48
    & 113.36 & 214.66 & 412.64
    & 115.17 & 222.07 & 434.49 \\
\hspace{1em} d1
    & 115.39 & 218.81 & 277.47
    & 123.25 & 237.25 & 404.29
    & 111.15 & 212.18 & 392.13
    & 115.36 & 225.80 & 437.36 \\
\hspace{1em} wd1
    & 114.83 & 225.94 & 321.35
    & 121.50 & 223.21 & 347.84
    & 84.52 & 87.63 & 91.93
    & 113.93 & 228.88 & 443.33 \\
\hspace{1em} GDPO
    & 114.39 & 213.37 & 252.97
    & 122.54 & 234.22 & 386.49
    & 115.04 & 217.39 & 339.80
    & 112.47 & 220.21 & 423.35 \\
\hspace{1em} DMPO
    & 106.20 & 130.93 & 157.57
    & 122.65 & 232.52 & 385.49
    & 112.77 & 214.88 & 346.28
    & 114.55 & 230.29 & 445.87 \\
\rowcolor[HTML]{F0F8FF}[\tabcolsep][\tabcolsep]
\rule[-0.85ex]{0pt}{3.1ex}\hspace{1em} PBPO (Ours)
    & 113.96 & 223.17 & 297.08
    & 118.92 & 224.55 & 363.46
    & 105.34 & 136.72 & 181.89
    & 113.63 & 220.81 & 438.93 \\
\bottomrule
\end{tabular}%
}
\end{table}

\paragraph{Coding benchmarks.}
Table~\ref{tab:code_avg_tokens} reports the corresponding average effective
sequence lengths on HumanEval and MBPP for both initialization checkpoints.

\begin{table}[t]
\centering
\caption{Average effective tokens across two coding benchmarks at generation
lengths 128, 256, and 512.}
\label{tab:code_avg_tokens}
\small
\setlength{\tabcolsep}{4pt}
\resizebox{\linewidth}{!}{%
\begin{tabular}{lcccccc}
\toprule
& \multicolumn{3}{c}{HumanEval}
& \multicolumn{3}{c}{MBPP} \\
\cmidrule(lr){2-4}
\cmidrule(lr){5-7}
Baseline / Seq Len
& 128 & 256 & 512
& 128 & 256 & 512 \\
\midrule
\textit{LLaDA-8B-Instruct}
    & 120.87 & 216.43 & 310.78
    & 118.14 & 206.96 & 265.30 \\
\hspace{1em} d1
    & 119.16 & 194.93 & 246.77
    & 114.72 & 189.01 & 241.53 \\
\hspace{1em} wd1
    & 112.15 & 144.70 & 165.38
    & 105.12 & 133.28 & 138.68 \\
\hspace{1em} GDPO
    & 116.87 & 223.54 & 393.85
    & 115.93 & 209.94 & 330.08 \\
\hspace{1em} DMPO
    & 116.68 & 206.83 & 284.57
    & 113.96 & 204.62 & 260.07 \\
\rowcolor[HTML]{F0F8FF}[\tabcolsep][\tabcolsep]
\rule[-0.85ex]{0pt}{3.1ex}\hspace{1em} PBPO (Ours)
    & 95.81 & 117.86 & 127.59
    & 84.72 & 93.35 & 97.53 \\
\midrule
\textit{LLaDA-1.5}
    & 120.82 & 220.32 & 317.95
    & 119.60 & 210.50 & 273.32 \\
\hspace{1em} d1
    & 118.32 & 207.38 & 269.42
    & 117.13 & 199.17 & 258.67 \\
\hspace{1em} wd1
    & 112.83 & 148.90 & 170.27
    & 104.30 & 127.45 & 145.92 \\
\hspace{1em} GDPO
    & 118.35 & 229.54 & 427.01
    & 118.48 & 222.93 & 384.49 \\
\hspace{1em} DMPO
    & 119.79 & 226.69 & 316.39
    & 119.39 & 214.48 & 275.40 \\
\rowcolor[HTML]{F0F8FF}[\tabcolsep][\tabcolsep]
\rule[-0.85ex]{0pt}{3.1ex}\hspace{1em} PBPO (Ours)
    & 111.04 & 144.02 & 172.57
    & 103.58 & 121.60 & 131.81 \\
\bottomrule
\end{tabular}%
}
\end{table}

\subsection{Robustness Across Various Evaluation Configurations}
\label{app:alternative_eval_settings}

We further evaluate LLaDA-1.5 and its RL-post-trained variants under five
additional decoding configurations. All configurations use a maximum
generation length of $256$ and $128$ denoising steps, while varying the block
length $B_l$ and token-selection strategy. For each configuration, we report
task accuracy and average effective sequence length across GSM8K, MATH500,
Countdown, Sudoku, HumanEval, and MBPP.

\paragraph{Low-confidence selection with $B_l=16$.}
Tables~\ref{tab:accuracy-g256-s128-bl16-low-confidence}
and~\ref{tab:effective-tokens-g256-s128-bl16-low-confidence} report accuracy
and average effective sequence length, respectively.

\begin{table}[t]
\centering
\caption{Accuracy (\%) across six benchmarks for LLaDA-1.5 baselines with generation length 256, 128 diffusion steps, block length 16, and low-confidence remasking. Best values in each benchmark column are in \textbf{bold}, and second-best distinct values are \underline{underlined}.}
\label{tab:accuracy-g256-s128-bl16-low-confidence}
\small
\setlength{\tabcolsep}{4pt}
\resizebox{1.0\linewidth}{!}{%
\begin{tabular}{lcccccc}
\toprule
\textbf{Baseline} & \textbf{GSM8K} & \textbf{MATH500} & \textbf{Countdown} & \textbf{Sudoku} & \textbf{HumanEval} & \textbf{MBPP} \\
\midrule
\textit{LLaDA-1.5}
    & 79.08 & 33.40 & 16.02 & 10.96 & 32.32 & 45.53 \\
\hspace{1em} d1
    & 80.21 & 35.40 & 19.53 & 18.66 & \underline{37.80} & \underline{47.08} \\
\hspace{1em} wd1
    & \underline{80.52} & 35.60 & 43.75 & 23.11 & 34.76 & 43.97 \\
\hspace{1em} GDPO
    & 79.76 & 36.00 & \underline{69.14} & 24.64 & 35.37 & \textbf{47.47} \\
\hspace{1em} DMPO
    & 47.69 & \underline{36.60} & 55.86 & \underline{25.02} & \textbf{41.46} & 44.75 \\
\rowcolor[HTML]{F0F8FF}[\tabcolsep][\tabcolsep]
\rule[-0.85ex]{0pt}{3.1ex}\hspace{1em} \textbf{PBPO (Ours)}
    & \textbf{84.38} & \textbf{38.40} & \textbf{73.83} & \textbf{38.09} & 37.20 & 45.53 \\
\bottomrule
\end{tabular}%
}
\end{table}

\begin{table}[t]
\centering
\caption{Average effective sequence length in tokens across six benchmarks for LLaDA-1.5 baselines with generation length 256, 128 diffusion steps, block length 16, and low-confidence remasking.}
\label{tab:effective-tokens-g256-s128-bl16-low-confidence}
\small
\setlength{\tabcolsep}{4pt}
\resizebox{1.0\linewidth}{!}{%
\begin{tabular}{lcccccc}
\toprule
\textbf{Baseline} & \textbf{GSM8K} & \textbf{MATH500} & \textbf{Countdown} & \textbf{Sudoku} & \textbf{HumanEval} & \textbf{MBPP} \\
\midrule
\textit{LLaDA-1.5}
    & 214.68 & 235.92 & 213.92 & 215.34 & 216.71 & 209.89 \\
\hspace{1em} d1
    & 218.30 & 236.67 & 209.41 & 224.58 & 203.32 & 199.69 \\
\hspace{1em} wd1
    & 225.21 & 223.35 & 87.38 & 231.79 & 150.53 & 132.07 \\
\hspace{1em} GDPO
    & 213.72 & 234.65 & 217.75 & 208.46 & 228.23 & 223.84 \\
\hspace{1em} DMPO
    & 136.23 & 233.55 & 214.05 & 229.50 & 226.51 & 218.67 \\
\rowcolor[HTML]{F0F8FF}[\tabcolsep][\tabcolsep]
\rule[-0.85ex]{0pt}{3.1ex}\hspace{1em} PBPO (Ours)
    & 222.81 & 224.18 & 147.56 & 219.10 & 145.73 & 121.81 \\
\bottomrule
\end{tabular}%
}
\end{table}

\paragraph{Low-confidence selection with $B_l=64$.}
Tables~\ref{tab:accuracy-g256-s128-bl64-low-confidence}
and~\ref{tab:effective-tokens-g256-s128-bl64-low-confidence} report accuracy
and average effective sequence length, respectively.

\begin{table}[t]
\centering
\caption{Accuracy (\%) across six benchmarks for LLaDA-1.5 baselines with generation length 256, 128 diffusion steps, block length 64, and low-confidence remasking. Best values in each benchmark column are in \textbf{bold}, and second-best distinct values are \underline{underlined}.}
\label{tab:accuracy-g256-s128-bl64-low-confidence}
\small
\setlength{\tabcolsep}{4pt}
\resizebox{1.0\linewidth}{!}{%
\begin{tabular}{lcccccc}
\toprule
\textbf{Baseline} & \textbf{GSM8K} & \textbf{MATH500} & \textbf{Countdown} & \textbf{Sudoku} & \textbf{HumanEval} & \textbf{MBPP} \\
\midrule
\textit{LLaDA-1.5}
    & \underline{80.44} & 35.80 & 21.48 & 7.70 & 32.32 & 43.97 \\
\hspace{1em} d1
    & 79.76 & 35.40 & 24.61 & 15.50 & 35.37 & 45.14 \\
\hspace{1em} wd1
    & 80.14 & 36.20 & 53.52 & 24.35 & 32.32 & 40.47 \\
\hspace{1em} GDPO
    & 78.17 & 33.20 & \textbf{69.92} & 25.17 & 34.15 & 44.75 \\
\hspace{1em} DMPO
    & 55.34 & \underline{36.60} & 60.16 & \underline{27.13} & \underline{37.80} & \underline{45.53} \\
\rowcolor[HTML]{F0F8FF}[\tabcolsep][\tabcolsep]
\rule[-0.85ex]{0pt}{3.1ex}\hspace{1em} \textbf{PBPO (Ours)}
    & \textbf{83.47} & \textbf{41.00} & \underline{67.58} & \textbf{40.33} & \textbf{40.24} & \textbf{48.64} \\
\bottomrule
\end{tabular}%
}
\end{table}

\begin{table}[t]
\centering
\caption{Average effective sequence length in tokens across six benchmarks for LLaDA-1.5 baselines with generation length 256, 128 diffusion steps, block length 64, and low-confidence remasking.}
\label{tab:effective-tokens-g256-s128-bl64-low-confidence}
\small
\setlength{\tabcolsep}{4pt}
\resizebox{1.0\linewidth}{!}{%
\begin{tabular}{lcccccc}
\toprule
\textbf{Baseline} & \textbf{GSM8K} & \textbf{MATH500} & \textbf{Countdown} & \textbf{Sudoku} & \textbf{HumanEval} & \textbf{MBPP} \\
\midrule
\textit{LLaDA-1.5}
    & 213.00 & 236.11 & 213.25 & 219.71 & 221.51 & 213.12 \\
\hspace{1em} d1
    & 218.31 & 236.60 & 211.13 & 226.96 & 211.35 & 201.00 \\
\hspace{1em} wd1
    & 226.24 & 224.47 & 80.70 & 229.16 & 145.35 & 120.46 \\
\hspace{1em} GDPO
    & 213.24 & 233.43 & 215.94 & 220.77 & 229.77 & 227.17 \\
\hspace{1em} DMPO
    & 129.48 & 234.84 & 215.18 & 229.09 & 225.58 & 211.99 \\
\rowcolor[HTML]{F0F8FF}[\tabcolsep][\tabcolsep]
\rule[-0.85ex]{0pt}{3.1ex}\hspace{1em} PBPO (Ours)
    & 223.12 & 226.33 & 123.18 & 222.70 & 145.60 & 121.19 \\
\bottomrule
\end{tabular}%
}
\end{table}

\paragraph{Random selection with $B_l=16$.}
Tables~\ref{tab:accuracy-g256-s128-bl16-random}
and~\ref{tab:effective-tokens-g256-s128-bl16-random} report accuracy and
average effective sequence length, respectively.

\begin{table}[t]
\centering
\caption{Accuracy (\%) across six benchmarks for LLaDA-1.5 baselines with generation length 256, 128 diffusion steps, block length 16, and random remasking. Best values in each benchmark column are in \textbf{bold}, and second-best distinct values are \underline{underlined}.}
\label{tab:accuracy-g256-s128-bl16-random}
\small
\setlength{\tabcolsep}{4pt}
\resizebox{1.0\linewidth}{!}{%
\begin{tabular}{lcccccc}
\toprule
\textbf{Baseline} & \textbf{GSM8K} & \textbf{MATH500} & \textbf{Countdown} & \textbf{Sudoku} & \textbf{HumanEval} & \textbf{MBPP} \\
\midrule
\textit{LLaDA-1.5}
    & 70.43 & 25.80 & 10.55 & 9.67 & 22.56 & 30.35 \\
\hspace{1em} d1
    & 74.07 & 31.00 & 16.80 & 16.08 & 24.39 & 34.24 \\
\hspace{1em} wd1
    & 74.45 & 27.80 & 35.16 & 23.68 & 23.17 & \underline{36.58} \\
\hspace{1em} GDPO
    & \underline{76.27} & 31.20 & \underline{57.42} & 23.30 & \textbf{29.27} & 36.19 \\
\hspace{1em} DMPO
    & 35.86 & \underline{31.80} & 44.92 & \underline{23.88} & \underline{26.83} & 31.91 \\
\rowcolor[HTML]{F0F8FF}[\tabcolsep][\tabcolsep]
\rule[-0.85ex]{0pt}{3.1ex}\hspace{1em} \textbf{PBPO (Ours)}
    & \textbf{79.38} & \textbf{33.00} & \textbf{64.06} & \textbf{36.60} & \textbf{29.27} & \textbf{37.74} \\
\bottomrule
\end{tabular}%
}
\end{table}

\begin{table}[t]
\centering
\caption{Average effective sequence length in tokens across six benchmarks for LLaDA-1.5 baselines with generation length 256, 128 diffusion steps, block length 16, and random remasking.}
\label{tab:effective-tokens-g256-s128-bl16-random}
\small
\setlength{\tabcolsep}{4pt}
\resizebox{1.0\linewidth}{!}{%
\begin{tabular}{lcccccc}
\toprule
\textbf{Baseline} & \textbf{GSM8K} & \textbf{MATH500} & \textbf{Countdown} & \textbf{Sudoku} & \textbf{HumanEval} & \textbf{MBPP} \\
\midrule
\textit{LLaDA-1.5}
    & 210.08 & 236.85 & 200.35 & 215.43 & 218.97 & 216.00 \\
\hspace{1em} d1
    & 213.42 & 233.93 & 188.89 & 213.16 & 208.24 & 192.67 \\
\hspace{1em} wd1
    & 225.39 & 220.64 & 80.82 & 227.92 & 144.90 & 131.38 \\
\hspace{1em} GDPO
    & 211.10 & 230.42 & 210.82 & 206.92 & 216.91 & 208.46 \\
\hspace{1em} DMPO
    & 223.87 & 229.37 & 206.87 & 216.55 & 219.80 & 214.87 \\
\rowcolor[HTML]{F0F8FF}[\tabcolsep][\tabcolsep]
\rule[-0.85ex]{0pt}{3.1ex}\hspace{1em} PBPO (Ours)
    & 222.92 & 221.82 & 164.72 & 220.76 & 148.30 & 128.40 \\
\bottomrule
\end{tabular}%
}
\end{table}

\paragraph{Random selection with $B_l=32$.}
Tables~\ref{tab:accuracy-g256-s128-bl32-random}
and~\ref{tab:effective-tokens-g256-s128-bl32-random} report accuracy and
average effective sequence length, respectively.

\begin{table}[t]
\centering
\caption{Accuracy (\%) across six benchmarks for LLaDA-1.5 baselines with generation length 256, 128 diffusion steps, block length 32, and random remasking. Best values in each benchmark column are in \textbf{bold}, and second-best distinct values are \underline{underlined}.}
\label{tab:accuracy-g256-s128-bl32-random}
\small
\setlength{\tabcolsep}{4pt}
\resizebox{1.0\linewidth}{!}{%
\begin{tabular}{lcccccc}
\toprule
\textbf{Baseline} & \textbf{GSM8K} & \textbf{MATH500} & \textbf{Countdown} & \textbf{Sudoku} & \textbf{HumanEval} & \textbf{MBPP} \\
\midrule
\textit{LLaDA-1.5}
    & 67.55 & 24.20 & 9.77 & 11.29 & 13.41 & 24.90 \\
\hspace{1em} d1
    & 70.13 & 25.60 & 18.36 & 15.79 & \textbf{19.51} & \underline{26.46} \\
\hspace{1em} wd1
    & \underline{73.84} & 24.80 & 37.11 & 22.87 & \textbf{19.51} & \textbf{28.40} \\
\hspace{1em} GDPO
    & 72.86 & 28.00 & \textbf{57.81} & \underline{24.78} & \textbf{19.51} & \textbf{28.40} \\
\hspace{1em} DMPO
    & 33.51 & \underline{28.80} & 45.70 & 24.11 & \underline{17.68} & 25.68 \\
\rowcolor[HTML]{F0F8FF}[\tabcolsep][\tabcolsep]
\rule[-0.85ex]{0pt}{3.1ex}\hspace{1em} \textbf{PBPO (Ours)}
    & \textbf{76.88} & \textbf{29.40} & \underline{50.39} & \textbf{35.74} & \textbf{19.51} & \underline{26.46} \\
\bottomrule
\end{tabular}%
}
\end{table}

\begin{table}[t]
\centering
\caption{Average effective sequence length in tokens across six benchmarks for LLaDA-1.5 baselines with generation length 256, 128 diffusion steps, block length 32, and random remasking.}
\label{tab:effective-tokens-g256-s128-bl32-random}
\small
\setlength{\tabcolsep}{4pt}
\resizebox{1.0\linewidth}{!}{%
\begin{tabular}{lcccccc}
\toprule
\textbf{Baseline} & \textbf{GSM8K} & \textbf{MATH500} & \textbf{Countdown} & \textbf{Sudoku} & \textbf{HumanEval} & \textbf{MBPP} \\
\midrule
\textit{LLaDA-1.5}
    & 208.34 & 236.18 & 194.70 & 216.06 & 221.29 & 212.11 \\
\hspace{1em} d1
    & 212.64 & 236.13 & 193.55 & 216.88 & 204.54 & 195.19 \\
\hspace{1em} wd1
    & 225.28 & 220.74 & 79.19 & 226.87 & 142.20 & 119.60 \\
\hspace{1em} GDPO
    & 211.06 & 233.40 & 206.23 & 204.16 & 219.20 & 211.19 \\
\hspace{1em} DMPO
    & 237.31 & 230.98 & 202.77 & 218.60 & 222.04 & 210.03 \\
\rowcolor[HTML]{F0F8FF}[\tabcolsep][\tabcolsep]
\rule[-0.85ex]{0pt}{3.1ex}\hspace{1em} PBPO (Ours)
    & 223.15 & 223.46 & 167.75 & 219.19 & 151.60 & 121.26 \\
\bottomrule
\end{tabular}%
}
\end{table}

\paragraph{Random selection with $B_l=64$.}
Tables~\ref{tab:accuracy-g256-s128-bl64-random}
and~\ref{tab:effective-tokens-g256-s128-bl64-random} report accuracy and
average effective sequence length, respectively.

\begin{table}[t]
\centering
\caption{Accuracy (\%) across six benchmarks for LLaDA-1.5 baselines with generation length 256, 128 diffusion steps, block length 64, and random remasking. Best values in each benchmark column are in \textbf{bold}, and second-best distinct values are \underline{underlined}.}
\label{tab:accuracy-g256-s128-bl64-random}
\small
\setlength{\tabcolsep}{4pt}
\resizebox{1.0\linewidth}{!}{%
\begin{tabular}{lcccccc}
\toprule
\textbf{Baseline} & \textbf{GSM8K} & \textbf{MATH500} & \textbf{Countdown} & \textbf{Sudoku} & \textbf{HumanEval} & \textbf{MBPP} \\
\midrule
\textit{LLaDA-1.5}
    & 63.99 & 22.80 & 8.59 & 11.53 & \underline{15.85} & 21.01 \\
\hspace{1em} d1
    & 66.26 & 25.80 & 20.31 & 17.42 & 14.02 & 20.62 \\
\hspace{1em} wd1
    & 68.31 & 25.40 & 35.55 & 23.92 & \textbf{17.68} & \underline{28.02} \\
\hspace{1em} GDPO
    & \underline{70.58} & \textbf{29.20} & \textbf{60.16} & 23.44 & \underline{15.85} & 20.23 \\
\hspace{1em} DMPO
    & 29.04 & \underline{27.60} & 45.70 & \underline{24.35} & 12.80 & 24.51 \\
\rowcolor[HTML]{F0F8FF}[\tabcolsep][\tabcolsep]
\rule[-0.85ex]{0pt}{3.1ex}\hspace{1em} \textbf{PBPO (Ours)}
    & \textbf{75.59} & 26.00 & \underline{51.56} & \textbf{36.22} & \underline{15.85} & \textbf{30.74} \\
\bottomrule
\end{tabular}%
}
\end{table}

\begin{table}[t]
\centering
\caption{Average effective sequence length in tokens across six benchmarks for LLaDA-1.5 baselines with generation length 256, 128 diffusion steps, block length 64, and random remasking.}
\label{tab:effective-tokens-g256-s128-bl64-random}
\small
\setlength{\tabcolsep}{4pt}
\resizebox{1.0\linewidth}{!}{%
\begin{tabular}{lcccccc}
\toprule
\textbf{Baseline} & \textbf{GSM8K} & \textbf{MATH500} & \textbf{Countdown} & \textbf{Sudoku} & \textbf{HumanEval} & \textbf{MBPP} \\
\midrule
\textit{LLaDA-1.5}
    & 207.31 & 232.23 & 189.29 & 210.69 & 221.56 & 213.26 \\
\hspace{1em} d1
    & 213.74 & 236.24 & 184.30 & 212.71 & 197.27 & 195.75 \\
\hspace{1em} wd1
    & 226.47 & 219.60 & 71.83 & 227.71 & 135.05 & 114.01 \\
\hspace{1em} GDPO
    & 212.44 & 234.97 & 197.49 & 202.65 & 222.63 & 216.20 \\
\hspace{1em} DMPO
    & 259.03 & 234.39 & 197.01 & 231.22 & 218.78 & 214.25 \\
\rowcolor[HTML]{F0F8FF}[\tabcolsep][\tabcolsep]
\rule[-0.85ex]{0pt}{3.1ex}\hspace{1em} PBPO (Ours)
    & 223.22 & 225.75 & 130.59 & 219.43 & 148.45 & 126.32 \\
\bottomrule
\end{tabular}%
}
\end{table}

\clearpage
\section{Proof for Theorem~\ref{thm:block_noise}}
\label{app:theorem-proof}

This section presents the proof for Theorem~\ref{thm:block_noise}. The comparison concerns the unclipped, advantage-free gradient
contributions displayed in the theorem. Before presenting the proof, we first
state the assumptions required for the variance comparison and clarify why they
are appropriate for this analysis.

\paragraph{Assumptions.}
Fix the prompt $q$, rollout index $i$, block index $b$, and prefix-block state
$z_i^{<b}$. We consider a full block with $B_l\geq 2$ and define its index set
as
\[
\mathcal{K}_b
=
\{(b-1)B_l+1,\ldots,bB_l\}.
\]
All expectations and covariances below are conditional on $(q,z_i^{<b})$ and
are taken under the factorized one-step behavior surrogate
\[
\prod_{k\in\mathcal{K}_b}
\pi_{\mathrm{old}}(o_i^k\mid q,z_i^{<b})
\]
used in Equation~\ref{eq:block_probability}. For each
$k\in\mathcal{K}_b$, abbreviate the token surrogate ratio and score vector by
\begin{equation}
\rho_i^k
=
\frac{
\pi_\theta(o_i^k\mid q,z_i^{<b})
}{
\pi_{\mathrm{old}}(o_i^k\mid q,z_i^{<b})
},
\qquad
s_i^k
=
\nabla_\theta\log\pi_\theta(o_i^k\mid q,z_i^{<b}).
\label{eq:refined_ratio_score}
\end{equation}
We assume that the pairs $\{(\rho_i^k,s_i^k)\}_{k\in\mathcal{K}_b}$ are
conditionally independent and identically distributed. We further assume that
$\pi_\theta$ and $\pi_{\mathrm{old}}$ have common positive support over the
finite vocabulary $\mathcal{V}$, that $\pi_\theta$ is differentiable with
parameter-independent support, and that the second moments used below are
finite.

Because $g_{\mathrm{block}}$ and $g_{\mathrm{token}}$ are vectors, we interpret
the variance in Theorem~\ref{thm:block_noise} as total variance:
\begin{equation}
\operatorname{Var}[g]
:=
\operatorname{tr}\!\left(\operatorname{Cov}(g)\right)
=
\mathbb{E}\!\left[
\left\|g-\mathbb{E}[g]\right\|_2^2
\right].
\label{eq:refined_total_variance}
\end{equation}
By the identical-distribution assumption, the moments
\begin{equation}
m_2
:=
\mathbb{E}\!\left[(\rho_i^k)^2\right],
\qquad
H
:=
\mathbb{E}\!\left[(\rho_i^k)^2s_i^k(s_i^k)^\top\right],
\qquad
c
:=
\mathbb{E}\!\left[(\rho_i^k)^2s_i^k\right]
\label{eq:refined_moments}
\end{equation}
do not depend on $k\in\mathcal{K}_b$. For the strict comparison, we additionally
assume that $\operatorname{Var}[\rho_i^k]>0$ and $H\neq0$.

\begin{remark}[Factorized surrogate assumptions]
The conditional independence assumption follows from the factorized one-step
block surrogate in Equation~\ref{eq:block_probability}. It is an assumption
about the surrogate probability space used in this comparison, not a claim
that tokens produced by the true multi-step denoising trajectory are
independent. Identical ratio-score moments across positions are an additional
homogeneity assumption that isolates the effect of importance-weighting
granularity.
\end{remark}

\begin{remark}[Regularity and strictness]
For standard full-softmax policies, all vocabulary tokens have positive
probability, so the common-support condition is satisfied. With a finite
vocabulary and finite score vectors, the required moments are also finite. The
positive-ratio-variance and nonzero-score conditions characterize a nontrivial
policy update and are necessary for strictness: if
$\pi_\theta=\pi_{\mathrm{old}}$, or if $B_l=1$, the two contributions can
coincide. A partial final block follows by replacing $B_l$ with its actual
number of positions, provided that this number is at least two.
\end{remark}

\begin{remark}[Advantage and clipping]
Theorem~\ref{thm:block_noise} compares the unclipped, advantage-free
contributions displayed in its statement. Multiplying both contributions by
the same fixed nonzero advantage $A_i$ scales their covariance matrices by
$A_i^2$ and preserves the strict comparison. PPO-style clipping introduces a
piecewise dependence on the surrogate ratios, so the covariance identities
below do not directly characterize the complete clipped-objective gradients.
\end{remark}

Having stated and justified the assumptions, we now prove the variance
comparison.

\begin{proof}
The common-support assumption and normalization of the token distributions give
\begin{align}
\mathbb{E}[\rho_i^k]
&=
\sum_{v\in\mathcal{V}}
\pi_{\mathrm{old}}(o_i^k=v\mid q,z_i^{<b})
\frac{
\pi_\theta(o_i^k=v\mid q,z_i^{<b})
}{
\pi_{\mathrm{old}}(o_i^k=v\mid q,z_i^{<b})
}
\\
&=
\sum_{v\in\mathcal{V}}
\pi_\theta(o_i^k=v\mid q,z_i^{<b})
=1.
\label{eq:refined_ratio_mean}
\end{align}
Differentiability and parameter-independent support similarly yield the
importance-weighted score identity
\begin{align}
\mathbb{E}[\rho_i^k s_i^k]
&=
\sum_{v\in\mathcal{V}}
\pi_\theta(o_i^k=v\mid q,z_i^{<b})
\nabla_\theta
\log\pi_\theta(o_i^k=v\mid q,z_i^{<b})
\\
&=
\nabla_\theta
\sum_{v\in\mathcal{V}}
\pi_\theta(o_i^k=v\mid q,z_i^{<b})
=0.
\label{eq:refined_weighted_score}
\end{align}
Equation~\ref{eq:refined_ratio_mean} also implies
\begin{equation}
m_2
=
1+\operatorname{Var}[\rho_i^k]
\geq 1.
\label{eq:refined_second_moment}
\end{equation}

Using Equation~\ref{eq:refined_ratio_score}, the token-level contribution in
Theorem~\ref{thm:block_noise} is
\[
g_{\mathrm{token}}
=
\sum_{k\in\mathcal{K}_b}\rho_i^k s_i^k.
\]
Equation~\ref{eq:refined_weighted_score} gives
$\mathbb{E}[g_{\mathrm{token}}]=0$. Moreover, conditional independence implies
that, for distinct $k,\ell\in\mathcal{K}_b$,
\[
\mathbb{E}\!\left[
\rho_i^k s_i^k(\rho_i^\ell s_i^\ell)^\top
\right]
=
\mathbb{E}[\rho_i^k s_i^k]
\mathbb{E}[\rho_i^\ell s_i^\ell]^\top
=0.
\]
Consequently, all cross-position terms vanish and
\begin{equation}
\operatorname{Cov}(g_{\mathrm{token}})
=
\sum_{k\in\mathcal{K}_b}
\mathbb{E}\!\left[
(\rho_i^k)^2s_i^k(s_i^k)^\top
\right]
=
B_l H.
\label{eq:refined_token_covariance}
\end{equation}

The block-level contribution can be written in the same notation as
\begin{align}
g_{\mathrm{block}}
&=
\left(
\prod_{j\in\mathcal{K}_b}\rho_i^j
\right)
\left(
\sum_{k\in\mathcal{K}_b}s_i^k
\right)
\\
&=
\sum_{k\in\mathcal{K}_b}
\rho_i^k s_i^k
\prod_{\substack{j\in\mathcal{K}_b\\j\neq k}}
\rho_i^j.
\label{eq:refined_block_expansion}
\end{align}
By conditional independence, together with
Equations~\ref{eq:refined_ratio_mean} and
\ref{eq:refined_weighted_score},
\begin{align}
\mathbb{E}[g_{\mathrm{block}}]
&=
\sum_{k\in\mathcal{K}_b}
\mathbb{E}[\rho_i^k s_i^k]
\prod_{\substack{j\in\mathcal{K}_b\\j\neq k}}
\mathbb{E}[\rho_i^j]
=0.
\label{eq:refined_block_mean}
\end{align}
Its covariance therefore equals its second moment. Expanding that moment and
separating the terms with matching and distinct token indices gives
\begin{align}
\operatorname{Cov}(g_{\mathrm{block}})
&=
\sum_{k\in\mathcal{K}_b}
\mathbb{E}\!\left[
(\rho_i^k)^2s_i^k(s_i^k)^\top
\prod_{\substack{j\in\mathcal{K}_b\\j\neq k}}
(\rho_i^j)^2
\right]
\\
&\quad+
\sum_{\substack{k,\ell\in\mathcal{K}_b\\k\neq\ell}}
\mathbb{E}\!\left[
(\rho_i^k)^2(\rho_i^\ell)^2s_i^k(s_i^\ell)^\top
\prod_{\substack{j\in\mathcal{K}_b\\j\neq k,\ell}}
(\rho_i^j)^2
\right]
\\
&=
B_l m_2^{B_l-1}H
+
B_l(B_l-1)m_2^{B_l-2}cc^\top.
\label{eq:refined_block_covariance}
\end{align}
Combining Equations~\ref{eq:refined_token_covariance} and
\ref{eq:refined_block_covariance} yields
\begin{align}
&\operatorname{Cov}(g_{\mathrm{block}})
-
\operatorname{Cov}(g_{\mathrm{token}})
\\
&\qquad=
B_l\left(m_2^{B_l-1}-1\right)H
+
B_l(B_l-1)m_2^{B_l-2}cc^\top.
\label{eq:refined_covariance_difference}
\end{align}
Both $H$ and $cc^\top$ are positive semidefinite. Since $m_2\geq1$ and
$B_l\geq2$, both coefficients on the right-hand side of
Equation~\ref{eq:refined_covariance_difference} are nonnegative. Thus,
\begin{equation}
\operatorname{Cov}(g_{\mathrm{block}})
\succeq
\operatorname{Cov}(g_{\mathrm{token}}).
\label{eq:refined_covariance_order}
\end{equation}

It remains to establish the strict scalar comparison. Since
$\operatorname{Var}[\rho_i^k]>0$,
Equation~\ref{eq:refined_second_moment} gives $m_2>1$. In
addition, the assumed nonzero importance-weighted score second moment implies
$H\neq0$; because $H\succeq0$, it follows that
$\operatorname{tr}(H)>0$. Taking the trace of
Equation~\ref{eq:refined_covariance_difference} therefore gives
\begin{align}
&\operatorname{Var}[g_{\mathrm{block}}]
-
\operatorname{Var}[g_{\mathrm{token}}]
\\
&\qquad=
B_l\left(m_2^{B_l-1}-1\right)\operatorname{tr}(H)
+
B_l(B_l-1)m_2^{B_l-2}\|c\|_2^2
>0.
\end{align}
By the total-variance convention in
Equation~\ref{eq:refined_total_variance}, this establishes the inequality in
Theorem~\ref{thm:block_noise} under the assumptions stated above.
\end{proof}

\section{Qualitative Examples}
\label{app:qualitative_examples}
\definecolor{ModelResponseBackground}{HTML}{F0F8FF}
\definecolor{ExampleHeaderBorder}{HTML}{B8C8CF}
\newtcolorbox{responseexample}[1][]{
    enhanced,
    breakable,
    colback=ModelResponseBackground,
    colframe=ExampleHeaderBorder,
    colbacktitle=ExampleHeaderBorder,
    coltitle=black,
    fonttitle=\small\raggedright,
    fontupper=\small,
    before upper=\raggedright,
#1
}

We present representative outputs from PBPO initialized from LLaDA-1.5 and LLaDA-8B-Instruct.
The responses are reproduced from the saved evaluation generations, with only
special end-of-sequence and padding tokens removed.

\subsection{MATH500 Examples}

\clearpage

\begin{responseexample}[title={\textbf{Model:} LLaDA-1.5 + PBPO\\
           \textbf{Dataset:} MATH500\\
           \textbf{Generation Sequence Length:} $512$\\
           \textbf{Question:} What is the probability that a randomly selected
           integer in the set $\{1,2,3,\ldots,100\}$ is divisible by 2 and not
           divisible by 3? Express your answer as a common fraction.}
]
\textbf{Model Response}

\texttt{<reasoning>}

First, we determine the total number of integers in the set
\(\{1, 2, 3, \ldots, 100\}\), which is 100.

Next, we count the number of integers divisible by 2 in this set. The sequence
of integers divisible by 2 is \(2, 4, 6, \ldots, 100\). This is an arithmetic
sequence with the first term \(a = 2\) and common difference \(d = 2\). The
number of terms in this sequence is found by solving
\[
a_n = a + (n-1)d \implies 100 = 2 + (n-1) \cdot 2
\implies 100 = 2n \implies n = 50.
\]
So, there are 50 integers divisible by 2.

Now, we count the number of integers divisible by both 2 and 3 (i.e., divisible
by 6). The sequence of integers divisible by 6 is \(6, 12, 18, \ldots, 96\).
This is an arithmetic sequence with the first term \(a = 6\) and common
difference \(d = 6\). The number of terms in this sequence is found by solving
\[
a_n = a + (n-1)d \implies 96 = 6 + (n-1) \cdot 6
\implies 96 = 6n \implies n = 16.
\]
So, there are 16 integers divisible by 6 in the set.

The number of integers divisible by 2 but not divisible by 3 is the total number
of integers divisible by 2 minus the number of integers divisible by 6:
\[
50 - 16 = 34.
\]

Thus, the probability that a randomly selected integer in the set
\(\{1, 2, 3, \ldots, 100\}\) is divisible by 2 and not divisible by 3 is
\[
\frac{34}{100} = \frac{17}{50}.
\]

\texttt{</reasoning>}

\texttt{<answer>}
\[
\boxed{\frac{17}{50}}
\]
\texttt{</answer>}
\end{responseexample}

\clearpage

\begin{responseexample}[title={\textbf{Model:} LLaDA-8B-Instruct + PBPO\\
           \textbf{Dataset:} MATH500\\
           \textbf{Generation Sequence Length:} $512$\\
           \textbf{Question:} What is the distance, in units, between the
           points $(2,-6)$ and $(-4,3)$? Express your answer in simplest
           radical form.}
]
\textbf{Model Response}

\texttt{<reasoning>}

To find the distance between two points \((x_1,y_1)\) and \((x_2,y_2)\), we use
 the distance formula:
\[
d=\sqrt{(x_2-x_1)^2+(y_2-y_1)^2}.
\]
Given the points \((2,-6)\) and \((-4,3)\), we substitute these coordinates
into the formula:
\[
d=\sqrt{((-4)-2)^2+(3-(-6))^2}.
\]
Simplify inside the parentheses:
\[
=-6 \quad \text{and} \quad 3+6=9.
\]
Square these results:
\[
(-6)^2=36 \quad \text{and} \quad 9^2=81.
\]
Add the squares:
\[
36+81=117.
\]
Take the square root:
\[
d=\sqrt{117}.
\]
Simplify \(\sqrt{117}\):
\[
\sqrt{117}=\sqrt{9\times13}=3\sqrt{13}.
\]
Thus, the distance between the points is \(\boxed{3\sqrt{13}}\).

\texttt{</reasoning>}

\texttt{<answer>}
\[
\boxed{3\sqrt{13}}
\]
\texttt{</answer>}
\end{responseexample}

\clearpage
\subsection{MBPP Examples}

\begin{responseexample}[title={\textbf{Model:} LLaDA-1.5 + PBPO\\
           \textbf{Dataset:} MBPP\\
           \textbf{Generation Sequence Length:} $512$\\
           \textbf{Question:} Write a Python function
           number is divisible by 11.\\
           \textbf{Tests:} \texttt{is\_Diff(12345) == False},
           \texttt{is\_Diff(1212112) == True}, and
           \texttt{is\_Diff(1212) == False}.}
]
\textbf{Model Response}

\texttt{<reasoning>}

To determine if a number is divisible by 11, we can use the simple divis for 11
rule. The rule states that a number is divisible by 11 if the difference between
the sum of the digits in the odd positions and the sum of the digits in the even
positions is either 0 or a multiple of 11.

\texttt{</reasoning>}

\texttt{<answer>}

{\small\ttfamily
\noindent def is\_Diff(number):\\
\hspace*{1em}\# Convert the number to a string to easily iterate over digits\\
\hspace*{1em}digits = str(number)\\
\hspace*{1em}\# Initialize sums for odd and even positions\\
\hspace*{1em}odd\_sum = 0\\
\hspace*{1em}even\_sum = 0\\
\hspace*{1em}\# Iterate over the digits of the number\\
\hspace*{1em}for i in range(len(digits)):\\
\hspace*{2em}if i \% 2 == 0:\\
\hspace*{3em}even\_sum += int(digits[i])\\
\hspace*{2em}else:\\
\hspace*{3em}odd\_sum += int(digits[i])\\
\hspace*{1em}\# Calculate the difference between the sums\\
\hspace*{1em}difference = odd\_sum - even\_sum\\
\hspace*{1em}\# Check if the difference is 0 or a multiple of 11\\
\hspace*{1em}return difference == 0 or difference \% 11 == 0
}

\texttt{</answer>}
\end{responseexample}
\clearpage
\begin{responseexample}[title={\textbf{Model:} LLaDA-8B-Instruct + PBPO\\
           \textbf{Dataset:} MBPP\\
           \textbf{Generation Sequence Length:} $512$\\
           \textbf{Question:} Write a function to flatten a list and sum all
           of its elements.\\
           \textbf{Tests:}
           \texttt{recursive\_list\_sum(([1,2,[3,4],[5,6]])) == 21},
           \texttt{recursive\_list\_sum(([7,10,[15,14],[19,41]])) == 106},
           and
           \texttt{recursive\_list\_sum(([10,20,[30,40],[50,60]])) == 210}.}
]
\textbf{Model Response}

\texttt{<reasoning>}

To flatten a list and sum all of its elements, we need to recursively traverse
 the list and accumulate the sum of its elements.

\texttt{</reasoning>}

\texttt{<answer>}

{\small\ttfamily
\noindent def recursive\_list\_sum(lst):\\
\hspace*{1em}total = 0\\
\hspace*{1em}for item in lst:\\
\hspace*{2em}if isinstance(item, list):\\
\hspace*{3em}total += recursive\_list\_sum(item)\\
\hspace*{2em}else:\\
\hspace*{3em}total += item\\
\hspace*{1em}return total
}

\texttt{</answer>}
\end{responseexample}

\end{document}